\documentclass{article}

     \PassOptionsToPackage{numbers, compress}{natbib}

\usepackage[preprint]{neurips_2020}

\usepackage[utf8]{inputenc} 
\usepackage[T1]{fontenc}    
\usepackage{url}            
\usepackage{booktabs}       
\usepackage{amsfonts}       
\usepackage{nicefrac}       
\usepackage{microtype}      

\usepackage[whole]{bxcjkjatype}

\usepackage{graphicx}
\usepackage{subfigure}
\usepackage{booktabs} 

\usepackage{bm}
\usepackage{amsmath,amssymb,amsfonts}
\usepackage{natbib}
\usepackage{mathrsfs}

\usepackage{multirow}
\usepackage{tikz}

\usepackage[rightcaption]{sidecap}
\usepackage{wrapfig}

\usepackage[version=3]{mhchem}
\usepackage{amsthm}
\newtheorem{theorem}{Theorem}
\newtheorem{proposition}[theorem]{Proposition}
\newtheorem{definition}[theorem]{Definition}

\usepackage{hyperref}

\usepackage{color}

\newcommand{\nint}{\mathbb{Z}_+}

\title{Weisfeiler-Lehman Embedding \\ for Molecular Graph Neural Networks}

\author{%
  Katsuhiko Ishiguro$^*$ \qquad Kenta Oono$^{*\dagger}$ \qquad Kohei Hayashi$^*$ \\
  $*$ Preferred Networks, Inc. \qquad \qquad $\dagger$ The University of Tokyo\\
  Tokyo, Japan\\
  \texttt{k.ishiguro.jp@ieee.org, oono@preferred.jp,
  hayashi.kohei@gmail.com}  \\
}
\begin{document}

\maketitle

\begin{abstract}
  A graph neural network (GNN) is a good choice for predicting the chemical properties of molecules. Compared with other deep networks, however, the current performance of a GNN is limited owing to the ``curse of depth.'' Inspired by long-established feature engineering in the field of chemistry, we expanded an atom representation using Weisfeiler-Lehman (WL) embedding, which is designed to capture local atomic patterns dominating the chemical properties of a molecule. 
  In terms of representability, we show WL embedding can replace the first two layers of ReLU GNN --- a normal embedding and a hidden GNN layer --- with a smaller weight norm. We then demonstrate that WL embedding consistently improves the empirical performance over multiple GNN architectures and several molecular graph datasets. 
\end{abstract}

\section{Introduction}
When finding new drugs or materials, supervised learning problems with structured data are frequently encountered. One of the most straightforward task settings is to predict the chemical properties (e.g., the toxicity) of a \emph{molecular graph}, which describes how atoms such as hydrogen and oxygen are connected through chemical bonds~\citep{Wu18MoleculeNet}. Despite a tremendous number of different molecules, their properties often solely depend on local atomic patterns. For example, specific local patterns called \emph{functional groups} in organic molecules are strong evidence for determining such chemical properties~\citep{smith2020march}. Traditionally, this type of local information is incorporated as handcrafted features~\citep{wiener1947structural,todeschini2009molecular}. 

During the past few years, numerous studies have shown the benefit of applying a graph neural network (GNN) for molecular tasks~\citep{Duvenaud15NIPS,Gilmer17ICML,coley2019graph,Yang19arXiv,Jin18ICML,DeCao_Kipf18ICML,Bradshaw19ICLR}. 
Currently, GNNs seem to be on track to achieve what convolutional neural networks (CNNs) accomplished for image tasks, nearly eliminating the need for handcrafted image features (e.g., \citep{krizhevsky2012imagenet}). However, recent theoretical and empirical results suggest a different outcome. Although the depth is an essential factor in representation learning~\citep{bengio2013representation}, a GNN incurs a ``curse of depth'', i.e., the performance degrades when we stack too many layers~\citep{Li18AAAI,Xu19ICLR,NT_Maehara19arXiv,Oono_Suzuki20ICLR}. 
How this problem can be overcome with a limited number of layers remains an important issue. 

\begin{figure}[t]
    \centering
    \includegraphics[width=140mm]{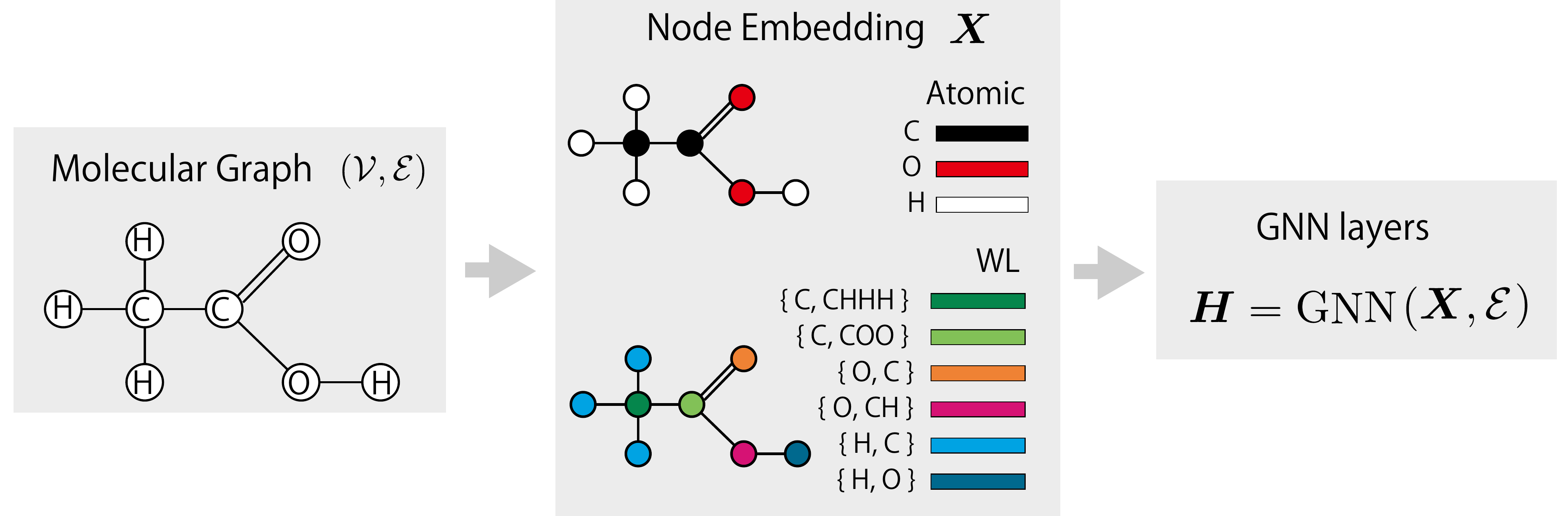}
    \caption{Overview of proposed WL embedding. 
    Left: A molecular graph. Each node is labeled by atoms such as carbon (C) and oxygen (O).
    Middle Upper: Atomic embedding constructs an embedding vector for each single atom. 
    Middle Lower: WL embedding constructs an embedding vectors for each combination of an atom and its neighboring atoms. 
    Right: The node embedding $\bm{X}$ is fed into a GNN with graph edges $\mathcal{E}$. 
    Note that hydrogen (H) atoms are not counted as nodes in practice but here we explicitly use them for illustrative purposes.  
    }
    \label{fig:WLE_overview}
\end{figure}

In this study, we revisit the idea of designing molecule features for a GNN. As a key contribution, we propose Weisfeiler-Lehman (WL) embedding, a simple, architecture-free embedding method that explicitly leverages the local atomic patterns. In contrast to a standard atomic embedding, which assigns a continuous vector to each single atom, WL embedding constructs an embedding vector by considering \emph{neighboring atoms}, from which we aim to extract subgraph information such as functional groups (Fig.~\ref{fig:WLE_overview}). 
Although WL embedding may generate too many embedding vectors and cause the overfitting problem, we can mitigate the problem by combining the normal embedding. 
In terms of representability, we show WL embedding is equivalent to the normal embedding with a single hidden ReLU layer while reducing the model complexity characterized by the weight norm, which is a key factor for better trainability. We also demonstrate that WL embedding improves the prediction performance regardless of the molecular datasets or GNN architectures applied. 

\paragraph{Notation}
Let $\nint = \{0\}\cup \mathbb{N}$ be the space of nonnegative integer. Let $G = (\mathcal{V}, \mathcal{E})$ be an undirected $K$-labeled multigraph where $\mathcal{V} = \{v_{i}\}_i$ is the set of nodes (i.e., atoms) and $\mathcal{E} = \{e_{i,j}\}_{i, j}$ is the multiset of edges (i.e., chemical bonds\footnote{Bonds can have multiple types (single, double, $\cdots$), but bond types are not used in the embedding. After embedding, GNNs may employ the bond type information~\citep{Schlichtkrull17arxiv,busbridge2019relational}.}) between nodes $v_i, v_j \in \mathcal{V}$. Each node $v_{i}$ is associated with a discrete label $\ell_{i}\in\{1,\dots, K\}$. We denote by $\mathcal{N}_{i}$ the neighboring nodes of $v_{i}$, where multiple edges are regarded as single edges. The empty set is denoted by $\emptyset$.

\section{Related Work}
\paragraph{GNNs and Curse of Depth}
After the seminal study of graph convolutional networks (GCNs) by \citep{Kipf_Welling17ICLR}, various GNN architectures have been proposed. For example, \citet{Gilmer17ICML} proposed a family of GNNs called message passing neural networks (MPNNs) and showed that MPNNs achieve the state-of-the-art performance for multiple molecular benchmark problems. Some researchers have proposed the use of GNNs specifically tailored for specific applications of molecular graphs~\citep{Bradshaw19ICLR,Jin19ICLR}. 
For general applications, a gated graph neural network (GGNN)~\citep{Li16ICLR} is a strong GRU-based GNN. 
The relational graph attention network (RelGAT)~\citep{busbridge2019relational} is a representative of a non-local neural network~\citep{Wang18CVPR}, which uses multiple attention mechanisms for a set of edge types. 

Along with their major successes, the limitations of GNNs have been gradually revealed. \citet{Li18AAAI} reported an \emph{over-smoothing} issue, that is, as the number of layers increases, the node representations become indistinguishable from each other. Although several studies~\citep{Li18AAAI,Xu19ICLR,li2019deepgcns,zhao2020pairnorm} have addressed this issue, no fundamental solution has yet been found, and particularly for molecular graphs, there are no practical deep architectures containing hundreds of layers, similar to ResNet~\citep{He16CVPR}.

\paragraph{WL Algorithm in Machine Learning}
WL algorithm~\citep{Weisfeiler_Lehman68} is a well-adopted heuristics to identify (non-)isomorphism between graphs. The essence of the algorithm is to expand a label of each node by concatenating labels of neighboring nodes and to compare these labels between graphs (see Supplementary material for the actual procedure). The WL graph kernels~\citep{Shervashidze09,Shervashidze11} adopted the idea of the WL label expansions directly as kernel features.
\citet{Togninalli19NeurIPS} applied the same idea to compute node representations for the graph kernels. Our study is on the same line, while we focus on GNNs instead of the graph kernels. We also extended the idea so that we can concurrently use the original labels to mitigate the sparsity problem; see the next section. 

In the context of GNNs, several researchers have incorporated the idea of the WL algorithm in building new layer architectures. 
For example, \citet{Jin17NIPS} and \citet{Lei17ICML} proposed MPNNs whose layer update rules are inspired through the mimicking of the WL kernel. 
\citet{Xu19ICLR} and \citet{Morris19AAAI} revealed that the WL algorithm has the upper bound capability of an MPNN, in terms of isomorphism testing. Based on the analysis, \citet{Xu19ICLR} proposed Graph Isomorphism Networks (GIN) where the function in each layer is modeled by MLP. 
\citet{Morris19AAAI} proposed another GNN architecture that can emulate the $k (>1)$-dimensional WL algorithm for further capability. 
In contrast to these studies addressing middle-layer architectures, we incorporated the label expansion operation into the node embedding layer, the cornerstone of GNNs. WL embedding is thus combined with any of the above architectures. 

Note that the theoretical upperbound of representaton powers of generic MPNNs does not eliminate the value of this work. 
Some fundamental theories such as the no-free-lunch theorem show the theoretical limitations of ML models for \textit{universal domains}.
These theorems motivated the research for models that leverage \textit{domain-specific} inductive bias to achieve empirical/theoretical performance in the domain.
Similarly, we focus on molecular graph analysis and show that 
a simple embedding inspired by ECFP (WL test) can improve generalization performance of molecular graph analysis, effective for multiple model-dataset pairs. 
We believe this is a concreate contribution in the domain-specific application of GNNs. 

\paragraph{Molecular Fingerprint}

Encoding graph-theoretical or chemical information of molecules into a symbolic representation, known as a molecular \text{fingerprint}, has been a central problem in computational chemistry, at least dating back to the 1940s~\citep{wiener1947structural}.
Fingerprints have a wide range of applications, such as a molecular similarity search or an analysis of quantitative structure-property relationships.

Owing to the long history of fingerprint in the area of computational chemistry, many methods have been proposed (c.f.,~\citealp{todeschini2009molecular}). 
Among them, extended connectivity fingerprint (ECFP)~\citep{rogers2010extended} is one of the most popular fingerprint approaches.
It is used to extract topological information of molecular graphs by enumerating and hashing the subgraphs of a graph within a specified radius. 
To achieve this, it aggregates the neighboring node information and expands the labels, similar to the WL algorithm.
In this sense, WL embedding shares a similar design concept with ECFP.
However, to the best of our knowledge, few attempts have been made to combine extended node labels to a GNN.
In addition, some variants of our method create node embeddings for the center and neighboring nodes separately, whereas ECFP does not.
We later show that this modification improves the empirical predictive performance for many different tasks.

\section{WL Embedding}
Node embedding is the first step of GNNs, which encodes a node $v_{i}$ into a continuous node feature vector $x_{i} \in \mathbb{R}^{d}$. Node embedding is implemented with $J\in\mathbb{N}$ embedding vectors $\{\theta_j\in\mathbb{R}^d\mid j=1,\dots,J\}$ and a hash function $s$ as
\begin{equation}\label{eq:x_i_embed_onehot}
    x_{i} = \text{Embed} \left( s(i) \right) = \theta_{ s(i) }.
\end{equation}
Usually, we assign the embedding vector by the node label $\ell_i$, i.e., $s(i)=\ell_i$, which we call \emph{atomic embedding}. 

To incorporate the neighbouring information, we use an extended label that is a tuple of a node label $\ell_i$ and its neighbouring labels $\mathcal{M}_i=\mathrm{multiset}\{\ell_j\mid j\in\mathcal{N}_i\}$\footnote{A multiset is a generalization of a set that can contain overlapping elements.} as a hash key. Namely, inspired by WL algorithm, we extend the hash function as $s(i) = t(\ell_{i}, \mathcal{M}_{i})$ where $t$ is an injective map from the extended label $(\ell_i, \mathcal{M}_i)$ to an integer $j\in\{1,\dots,J\}$. Now we introduce WL embedding as
\begin{equation}
    \textbf{(Naive) WL Embed:} \quad 
    x_{i} 
    = \text{WLEmbed} \left(\ell_i\, , \mathcal{M}_i \right) 
    = \text{Embed} \left( t(\ell_{i}\, , \mathcal{M}_{i}) \right).
    \label{eq:naive_WL_embed}
\end{equation}

Fig.~\ref{fig:WLE_overview} illustrates how WL embedding works. In contrast to atomic embedding, WL embedding makes every embedded vector of carbon and oxygen atoms distinguishable because its neighborhoods are all different. This allows GNNs to treat frequent atoms that appeared in molecules in a more detailed way. 

\subsection{Extensions}\label{sec:c_wl_g_wl}

Although the naive form of WL embedding~\eqref{eq:naive_WL_embed} increases the representability of atoms, it may cause a sparsity problem. Namely, it may produce many extended labels that are rarely used, and learning the embedding vectors involving less-frequent atoms such as sulfur becomes more challenging with limited data.

In this paper, we propose the extensions of the naive WL idea that allows more robust and memory-efficient node embedding for molecular graph analysis. 
Our solution is to separate the atomic and neighbouring information into two embedding vectors $z_{i,\ell}=\mathrm{WLEmbed}(\ell_i,\emptyset)$ and $z_{i,\mathcal{M}}=\mathrm{WLEmbed}(\emptyset,\mathcal{M}_i)$. Here, $z_{i,\ell}$ represents the atomic information, and, for example, every carbon node can share their knowledge through it. In contrast, $z_{i,\mathcal{M}}$ reflects the neighboring information, which can maintain the high representability. To implement this idea, we consider two approaches.

\paragraph{C-WL Embedding}

The first approach is to combine the two embedding vectors by concatenation. We define the concatenated WL (C-WL) embedding as follows:
\begin{alignat}{2}
    x_{i} &= W \cdot \text{Concat}\left[ z_{i,\ell} \, , z_{i,\mathcal{M}} \right] \, \in \mathbb{R}^{d} 
    \label{eq:CWL_embed}\, , \\
    z_{i,\ell} &= \text{WLEmbed}_{1} (\ell_{i},\emptyset) \in \mathbb{R}^{d_{1}} \, , 
    \label{eq:CWL_embed_z_il} \\
    z_{i,\mathcal{M}} &= \text{WLEmbed}_{2} (\emptyset, \mathcal{M}_{i}) \in \mathbb{R}^{d_{2}} \, , 
    \label{eq:CWL_embed_z_im} 
\end{alignat}
where $W \in \mathbb{R}^{d \times (d_{1} + d_{2})}$ is a trainable weight matrix for linearly mixing $z_{i, \ell}$ and $z_{i, \mathcal{M}}$.

\paragraph{G-WL Embedding}

Another approach is to employ a weighted interpolation between $z_{i,\ell}$ and $z_{i,\mathcal{M}}$, where the mixing weight is computed through a gate function. 
We formalize the above idea as the Gated-sum WL (G-WL) embedding defined by 
\begin{alignat}{2}
    x_{i} =& \left(1 - G_{i} \right) \odot z_{i,\ell}
    + G_{i} \odot z_{i,\mathcal{M}} \, \in \mathbb{R}^{d}  \, , 
    \label{eq:GWL_embed} \\
    G_{i} =& \sigma \left( 
    W_{1} z_{i,\ell} + W_{2} z_{i,\mathcal{M}}
    \right) \in [0,1]^d \, ,
    \label{eq:GWL_gate_func} \\
    z_{i,\ell} =& \text{WLEmbed} (\ell_{i}, \emptyset) \in \mathbb{R}^{d} \, , 
    \label{eq:GWL_embed_z_il} \\
    z_{i,\mathcal{M}} =& \text{WLEmbed} (\emptyset, \mathcal{M}_{i}) \in \mathbb{R}^{d} \, , 
    \label{eq:GWL_embed_z_im} 
\end{alignat}
where $\odot$ denotes the Hadamard product, $\sigma$ is the element-wise sigmoid function, and $W_1$, $W_2\in \mathbb{R}^{d\times d}$ are trainable parameters of gate functions $G_i$. 

The advantages of G-WL against naive WL are essentially the same as those of C-WL. 
As one difference, G-WL adopts an adaptive gate function for mixing $z_{i,\ell}$ and $z_{i,\mathcal{M}}$, whereas C-WL employs a fixed weight matrix to concatenate them. 
Adaptive gates allow the embedding module to tune the interpolated node embedding $x_{i}$ according to the combination of $z_{i,\ell}$ and $z_{i,\mathcal{M}}$. By contrast, parameters of the gate function require more complications of the training. 

\subsection{Combining with GNNs}

After obtaining the node embedding $\bm{X} = \{ x_{1},\dots,x_{|\mathcal{V}|} \}$, we feed them into multiple GNN layers such as MPNN layers. For example, an $L$-layered MPNN iteratively updates the node latent vectors $h_{i}^{(l)} \in \mathbb{R}^{d_l}$ for each layer $l=1,\dots,L$, where
\begin{equation}
    h_{i}^{(l)} = \text{UPD} \left( h_{i}^{(l-1)}, 
    \text{AGG}\left(\{ h_{j}^{(l-1)} \mid j \in \mathcal{N}_{i} \}\right)
    \right). \,  
    \label{eq:MPNN_h}
\end{equation}
Here, UPD(ate) and AGG(regate) are some appropriate functions~\citep{Gilmer17ICML}. 
In addition, $h_{i}^{(0)}$ is initialized using the node embedding $x_i$. The final updates of the latent vectors 
are denoted as $\bm{H} = \{h_{i}\}_{i} = \{h_{i}^{(L)}\}_{i}$. To predict a discrete or continuous target variable $y$, we aggregate $\bm{H}$ with an appropriate readout function and make prediction as $\hat{y} = \text{READOUT}(\bm{H})$.\footnote{Several authors proposed readout functions that additionally use graph topology (e.g., \cite{Ying18NeurIPS}). For simplicity, we do not consider such functions.} We train GNN layers, the readout function, and the embedding vectors. 

\subsection{Iterative Label Expansion}\label{sec:iterative_label_expansion}
As with the WL algorithm, we can further extend the labels by looking up more than one-hop neighbors (e.g., neighborhoods of neighborhoods). 
Iterating the extensions increases the representation power because expanded labels can exploit long-distant nodes.
By contrast, because the possible expanded label patterns combinatorially increase, too many extensions make the embeddings extremely sparse.
In Supplemental material, we empirically investigated 
this trade-off using real datasets. Iterating label expansion twice improves the performance of C-WL and G-WL compared to the single iteration case, when a dataset has a large sample size. On the other hand, more iterations deteriorate 
the performance of the naive WL for all datasets. 

\section{Analysis}
The process of WL embedding, generating node representations by gathering neighboring information, looks very similar to what the single iteration of GNN update rule~\eqref{eq:MPNN_h} does. So how are they different? How does it make a difference in the outcome?

To answer this question, first we investigate the representability of WL embedding. Let $\mathcal{G}$ be a finite set of graphs with $K$ node labels and $f:\mathcal{G}\to\{+1,-1\}$ be a GNN with a linear classifier $f(G) = \mathrm{sign}(\langle w', \bar{h}(G) \rangle + b')$ and sum readout layer $\bar{h}(G) = \sum_{i\in\mathcal{V}} h_i$ where  $h_i$ is the representation of node $i$ created either by a GNN or WL embedding. Since the dimensionality of the input (representation) space roughly determines the capacity of a linear classifier (e.g., VC dimension; see book~\cite{mohri2018foundations}), we check the best-possible dimensionality of a vector space spanned by $\bar{h}$. 
\begin{definition}
Let $\Theta$ be the set of GNN parameters including embedding vectors. We define the \emph{maximum dimensionality} of a readout output $\bar{h}_\theta$ for a graph set $\mathcal{G}$ by $\mathrm{MD}_{ \mathcal{G}}(\bar{h})=\sup_{\theta\in\Theta}\dim(\mathrm{span}\{\bar{h}_\theta(G)\mid G\in\mathcal{G}\})$.
\end{definition}
Suppose we employ the naive WL embedding vector~\eqref{eq:naive_WL_embed} as the node representation $h_i$. Since $h_i$ is completely determined by its label $\ell_i$ and neighbouring labels $\mathcal{M}_i$, we can enumerate their patterns. Given a graph $G\in\mathcal{G}$, let $\mathbf{m}=(m_1,\dots,m_K)$ be the multiplicities of neighbouring labels where $m_k\in\nint$ counts the number of connected nodes labeled as $k=1,\dots,K$. Let $n_{k,\mathbf{m}}\in\nint$ be the number of nodes labeled as $k$ with neighbourhood pattern $\mathbf{m}$. The readout is then written as
\begin{align}\label{eq:wle-lattice}
\bar{h}^{\mathrm{WLE}}(G) = \sum_{k=1}^K\sum_{\mathbf{m}} n_{k, \mathbf{m}} \text{WLEmbed} (k, \mathbf{m} )
\end{align}
(we interchangeably used the multiplicities $\mathbf{m}$ as the multiset $\mathcal{M}$.) We see that $\bar{h}^{\mathrm{WLE}}$ is in the linear space spanned by the embedding vectors, and its degree of freedom is maximized when all the embedding vectors are linearly independent. WL embedding immediately achieves this because by definition it can assign a unique vector for each pair $(k, \mathbf{m} )$. When any node in $\mathcal{G}$ connects with at most $M$ nodes of the same label (i.e., $m_k\leq M$ for $k=1,\dots,K$), the number of the combinations of the pairs is bounded above by $K(M+1)^K$, which determines the maximum dimensionality.
\begin{proposition}
Let $\mathcal{G}_{K, M}$ be the set of graphs with $K$ node labels and label-specific maximum degree $M$. The maximu dimensionality of WL embedding~\eqref{eq:wle-lattice} is bounded as $\mathrm{MD}_{\mathcal{G}_{K, M}}(\bar{h}^{\mathrm{WLE}})\leq K(M+1)^K$. 
The upper bound is achieved if the embedding dimension $d$ satisfies $d \geq K(M+1)^K$. 
\end{proposition}

Next, suppose the node representation $h_i$ is given by a single GNN layer defined as
\begin{align}\label{eq:nonlinear-gnn}
    h_i &= \sigma(U_{\ell_i} x_i + W_{\ell_i}\sum_{j \in \mathcal{N}_i} V_{\ell_i \ell_j} x_j)
\end{align}
where $\sigma:\mathbb{R}^{d_1}\to\mathbb{R}^{d_1}$ is an activation function, $\{V_{k l}\in\mathbb{R}^{d\times d}, U_{k} \in \mathbb{R}^{d_1\times d}, W_k \in\mathbb{R}^{d_1\times d_1} \mid k,l = 1,\dots, K \}$ are label-specific weight matrices, and $x_i = \mathrm{Embed}(\ell_i)\in\mathbb{R}^d$ is the atomic embedding vector of node $i$. GNNs defined by \eqref{eq:nonlinear-gnn} can be seen as a restricted class of multiset-broadcasting GNNs~\cite{sato2019approximation}, including GraphSAGE~\cite{Hamilton17NIPS}, GCN~\cite{Kipf_Welling17ICLR},  
and GIN~\cite{Xu19ICLR}. Using the same notation of \eqref{eq:wle-lattice}, the readout output is written as
\begin{align}\label{eq:relu_gnn}
    \bar{h}^{\mathrm{GNN}}(G) &= \sum_{k=1}^K \sum_{\mathbf{m}} n_{k,\mathbf{m}} \sigma( W_k L_k(\mathbf{m}) + b_k)
\end{align}
where $b_k = U_k x_k \in\mathbb{R}^{d_1}$, $L_k(\mathbf{m}) = \sum_{l=1}^K m_l e_{kl} \in\mathbb{R}^{d}$, and $e_{kl}=V_{kl}x_l \in\mathbb{R}^{d}$ for $k,l=1,\dots,K$. Here, $L_k(\mathbf{m})$ is a point of the $K$-dimensional lattice (grid) spanned by the basis $\{e_{kl} \mid l=1,\dots,K\}$. To maximize the dimensionality of $\bar{h}^{\mathrm{GNN}}$, the activation function $\sigma$ needs to map every grid point of $L_k(\cdot)$ to a linearly independent vector. 
Following the strategy of \citet{yun2018small}, we show the construction of such mapping with the rectifier activation. The proof is deferred to Supplementary material.

\begin{theorem}\label{thm:relu}
The maximum dimensionality of GNN~\eqref{eq:relu_gnn} is bounded as $\mathrm{MD}_{\mathcal{G}_{K,M}}(\bar{h}^{\mathrm{GNN}}) \leq K(M+1)^K$. With the rectifier activation $\sigma(x) = \max(x, 0)$, the upper bound is achieved if the embedding dimension $d$ and the layer width $d_1$ satisfy  $d \geq K, d_1 \geq K (M + 1)^K$ and the norm of the parameters satisfies $\max_l\|W_{k} V_{kl} x_l\|_2  = \Omega(M^{3K/2}), \|U_k x_k\|_2=\Omega(M^{3K/2})$ for $k=1,\dots,K$.
\end{theorem}

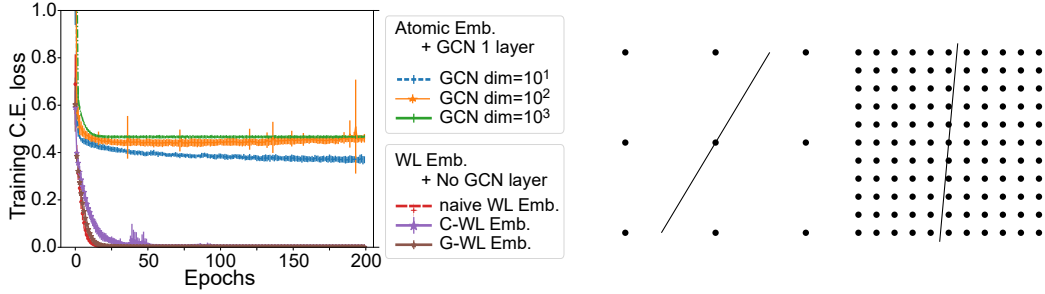
\begin{figure}
\raisebox{-0.5\height}{
  \includegraphics[width=76mm]{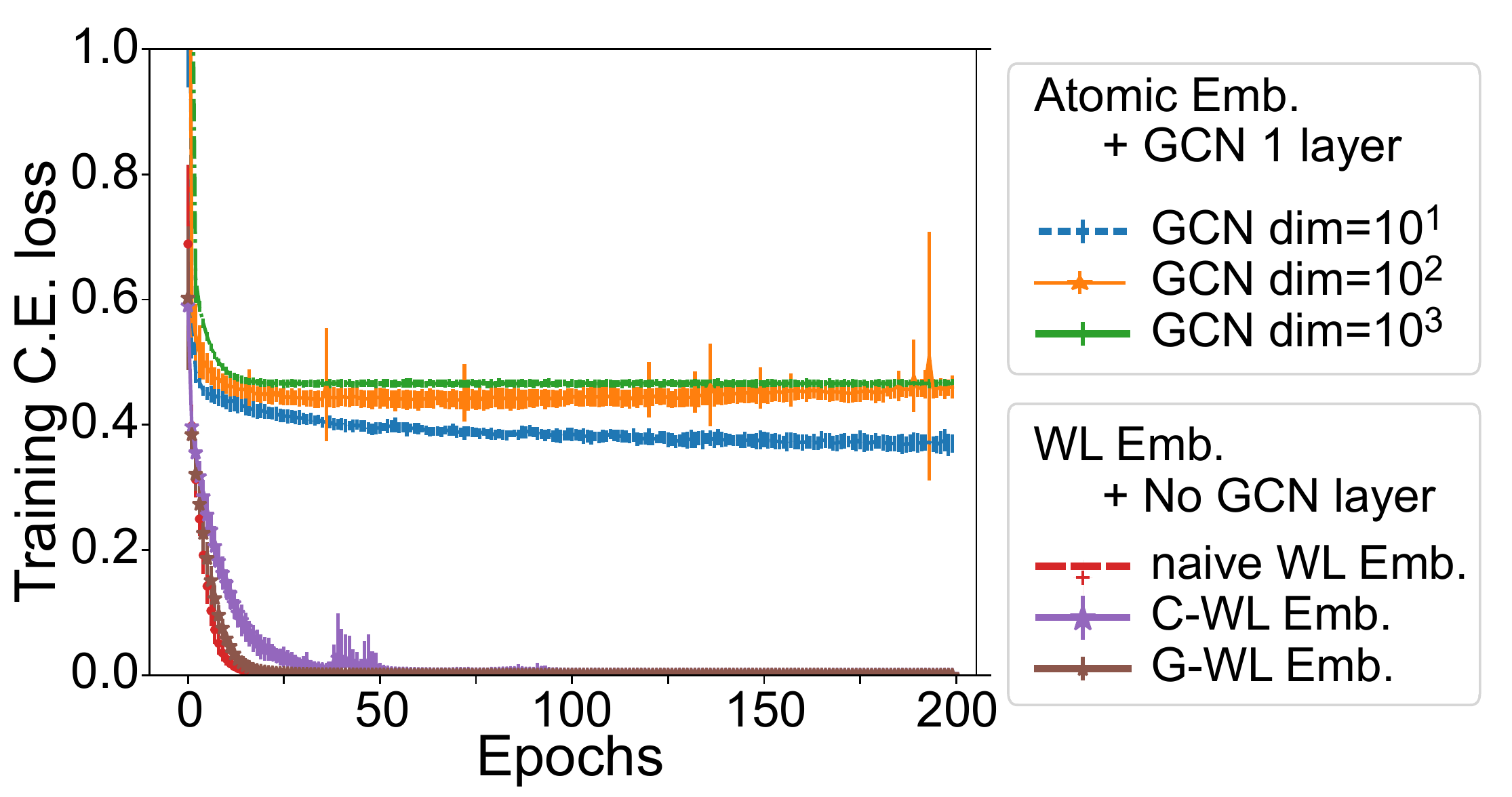}}
  \hspace{1em}
  \raisebox{-0.5\height}{
  \begin{tikzpicture}[scale=1.2]
    \foreach \i in {0,...,2}
      \foreach \j in {0,...,2}{
        \fill[black] (\i,\j) circle(1pt);
      };
      \draw (0.4, 0) -- (1.6, 2);
  \end{tikzpicture}}
  \hspace{1em}
  \raisebox{-0.5\height}{
  \begin{tikzpicture}[scale=1.2]
    \foreach \i in {0,0.2,...,2}
      \foreach \j in {0,0.2,...,2}{
        \fill[black] (\i,\j) circle(1pt);
      };
      \draw (0.9, -0.1) -- (1.1, 2.1);
  \end{tikzpicture}}
    
    \caption{Left: Learning curves in a synthetic experiment (the subgraph detection task in Section~\ref{sec:experiment_synthetic_graphs}). The atomic embedding is followed by a single layer of ReLU GCN with different layer width. WL embeddings are directly fed into the readout layer (no GCN layer). Right: 2D grids where $M=2$ and $M=10$ with hyperplanes (lines) that contain the center points while keeping away from the other points.}
    \label{fig:subgraph_detection_vs_dims_loss}
\end{figure}

Although Theorem~\ref{thm:relu} shows that WL embedding and the single ReLU layer have the same representation power, the condition of the parameter scale implies the realization is not efficient. The scale of the parameters\footnote{The scale increases with the number of labels $K$ and the label-specific degree $M$ because of the following reason. 
To assign an independent vector to each grid point, we need a layer unit that activates only by a specific point. We can create such a unit if there exists a hyperplane that only contains the point.
Although we can always find such hyperplanes, their gradient is inevitably increased as the number of points increases to avoid hitting the other grid points, as shown in Figure~\ref{fig:subgraph_detection_vs_dims_loss} (right).} $\Omega(M^{3K/2})$ is significantly larger than in a normal setting\footnote{For example, $\Theta(1/\sqrt{K})$ is a popular choice as the initial scale for ReLU layers (e.g. \cite{he2015delving}).}. Attaining such a large-scale solution is challenging for stochastic gradient descent and its variants because their implicit bias favors small-norm solutions~\cite{zhang2016understanding,soudry2017the}. Indeed, we empirically verified this. 
Figure~\ref{fig:subgraph_detection_vs_dims_loss} (left) shows that the atomic embedding with a single-layer GCN could not achieve the zero training error, showing the hardness of training (see the next section for the detailed setting).

\section{Experiment: Synthetic Graphs}\label{sec:experiment_synthetic_graphs}
To examine whether WL embedding behaves as we intended, we prepare two toy tasks: \emph{label counting} and \emph{subgraph detection}. The label counting task is to predict the number of specific node labels, which is solvable without graph structure. The subgraph detection task is to classify whether a graph has specific subgraphs, which requires neighboring information.

We prepare datasets consisting of two types of artificial graphs: \emph{positive} graphs and \emph{negative} graphs.
For the positive graphs, we first generate a $5$-node regular graph of degree $4$ and remove edges independently and randomly with probability $0.25$.
Then we attach one of three target subgraphs, all of which have $5$ nodes, by adding an edge to each node pair of two graphs independently randomly with probability $0.1$.
We adopt the generated graph if and only if it is connected, the degrees of all nodes are smaller than or equal to $4$, and it contains exactly one of the target subgraphs.
For the negative graphs, we generate a $10$-node regular graph of degree $4$ and remove edges independently randomly with probability $0.25$.
We adopt the generated graph if and only if it is connected and does not include any of the target subgraphs.
For both procedures, we associate one of five labels with each node of the graph independently uniformly randomly.

For each task, we generate three datasets. Each dataset has $300$ positive graphs and $300$ negative graphs.
For each pair of datasets and tasks, we conduct five runs with random initialization. All hyperparameters are manually fixed and shared in all trials. We adopt $L$-layer nonlinear GCNs with atomic and WL embeddings for $L=1,\dots,6$.

\begin{SCfigure}[10][t]
    \centering
    \includegraphics[width=45mm]{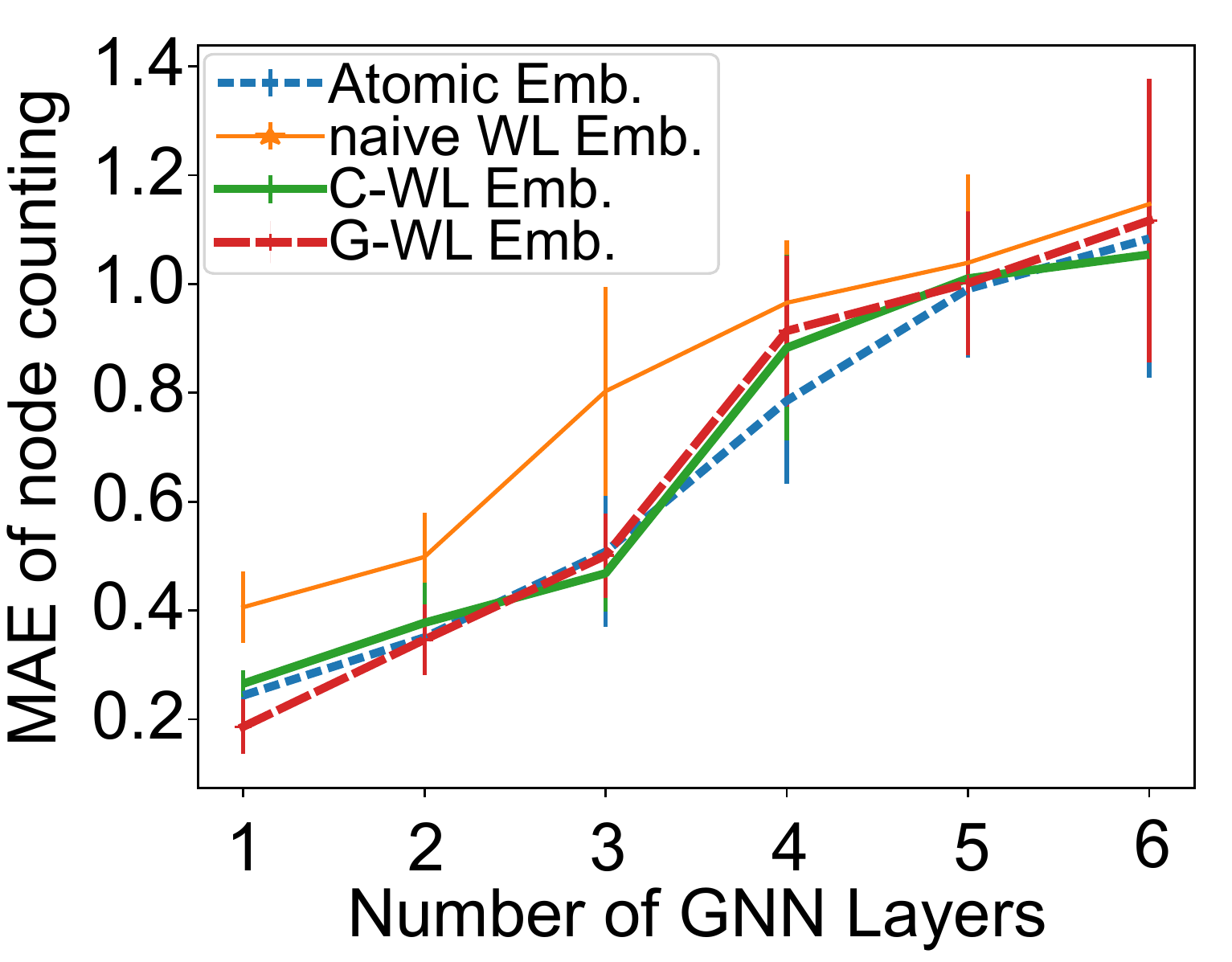}
    \includegraphics[width=45mm]{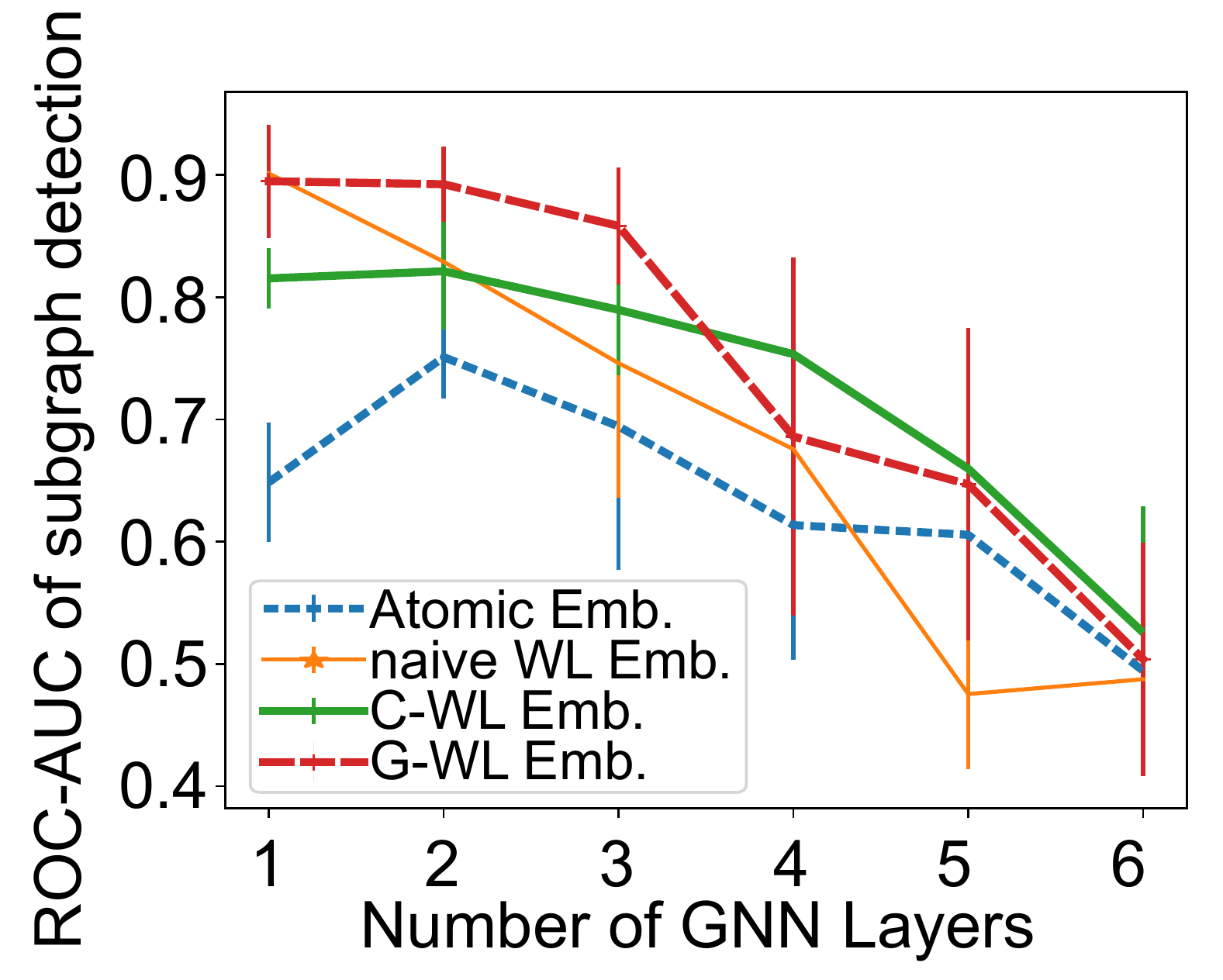}
    \caption{Experiments with artificial graph datasets. Averages and standard deviations of 15 trials are presented. 
    Left: MAEs (smaller is better) of the counting tasks over the number of layers. 
    Right: ROC-AUCs (larger is better) of the detection task over the number of layers. 
    }
    \label{fig:toy_problem_alpha0.01}
\end{SCfigure}

Figure~\ref{fig:toy_problem_alpha0.01} clearly shows that, while both embeddings behave mostly identically at the label counting task (Left), WL embeddings significantly outperform atomic embedding at the subgraph detection task (Right). This indicates that local structural information is actually important for node embedding, even with nonlinear activation. The results also capture the ``curse of depth''. At both tasks, the performance gradually declines after two layers. Nevertheless, WL embeddings always improve the performance at the detection tasks, implying that WL embeddings somehow dispel the curse for neighborhood-sensitive tasks.

\section{Experiment: Molecular Benchmarks}
In this section, we report our main empirical results on benchmark datasets. 
We compare the performances of the prediction among the aforementioned embedding techniques, combined with multiple GNN architectures. Details can be found in the supplementary material. 

\subsection{Datasets and Tasks}

We use eight datasets in MoleculeNet~\citep{Wu18MoleculeNet}: 
three for regression and five for classification. In all datasets, we used the train/validation/test data splits of a ``scaffold'' type, which is considered to be difficult for test predictions~\citep{Ruddigkeit12QM9,Ramakrishnan14QM9}.
For the graph regression tasks, we use the QM9 and the QM8 datasets from the field of \textit{quantum chemistry}, and the Lipophilicity (LIPO) dataset from the field of \textit{physical chemistry}. 
The task is to predict the associated numerical value(s) from the molecular graphs. 
We evaluate the performance of the models using the mean absolute errors (MAEs) over properties and molecules. 
For the graph classification tasks, we use the Tox21, the ToxCast and the Clintox datasets from the field of \textit{physiology}, and the HIV and the MUV datasets from the field of \textit{biophysics}. 
The task is to predict the binary label(s) from the molecular graphs. We evaluate the performance of the models using the ROC-AUC values over targets and molecules.

\subsection{Setting}\label{sec:experiment_setting}

We tested the embedding approaches on five popular GNN models: GCN, GGNN, RelGAT, GIN, and Neural FingerPrint (NFP)~\citep{Duvenaud15NIPS}, which is a GNN-based learnable molecular fingerprint. 
For each GNN, we implemented the readout function described in the original paper. We employed the softmax cross-entropy loss for classification tasks and the squared loss for regression tasks.
For each dataset-model pair, we optimized the following hyperparameters by using Optuna~\citep{Akiba19SIGKDD} based on the validation score: the number of GNN layers $L$, the dimension of latent hidden vectors and embedding vectors $d$, and the $\alpha$ of the Adam optimizer~\citep{Kingma_Ba15ICLR}. 
The other architecture-specific hyperparameters such as the MLP architecture of GIN layers were fixed throughout the study.
All the experiments were repeatedly conducted with 10 different random seeds. 

\subsection{Main Results}

Tables~\ref{tab:optimized_result_regression}~and~\ref{tab:optimized_result_classification} show the generalization performance (i.e., test scores) over $8$ (datasets) $\times$ $5$ (GNNs) $= 40$ cases. 
Overall, at least one of the WL embedding approaches consistently improves the generalization performance in all the cases except the Tox21-NFP pair. In particular, C-WL embedding performs better than the atomic embedding in 37 out of 40 cases. 
G-WL embedding and the naive WL embedding also perform better than the atomic embedding in many cases but are less stable compared to the C-WL. 
In addition, C-WL achieves the best test scores among four embeddings in 20 cases.
These results indicate the importance of incorporating neighboring node labels $\mathcal{M}$ into the node embedding for GNNs. 

\begin{table*}[t]
    \centering
    \small
    \begin{tabular}{cc||c@{\hskip 0.5\tabcolsep}c@{\hskip 0.5\tabcolsep}c@{\hskip 0.5\tabcolsep}c@{\hskip 0.5\tabcolsep}c}
         Dataset & Embedding & GCN &
                               RelGAT &
                               GGNN &
                               GIN &
                               NFP \\
        \hline 
        \multirow{4}{*}{LIPO} & Atomic    & 0.7077 &
                                            0.6298 &
                                            0.6670 & 
                                            0.7034 &
                                            0.7067 \\
                               & Naive WL &  \textbf{\qquad \textit{+0.0744}} & 
                                \qquad \textbf{+0.0462}  & 
                                \qquad -0.0716           & 
                                \qquad \textbf{\textit {+0.0641}} &
                                \qquad \textbf{+0.0582} \\
                               & C-WL     & \textbf{\qquad +0.0553}  & 
                                \qquad \textbf{\textit{+0.0487}} & 
                                \qquad \textbf{\textit{+0.0707}} & 
                                \qquad \textbf{+0.0179} & 
                                \qquad \textbf{\textit{+0.0713}} \\
                               & G-WL     & \textbf{\qquad +0.0688}  & 
                                \qquad \textbf{+0.0453}  & 
                                \qquad \textbf{+0.0255}  & 
                                \qquad \textbf{+0.0192} & 
                                \qquad \textbf{+0.0582}  \\
        \hline
        \multirow{4}{*}{QM9}   & Atomic   & 23.7247 &
                                            11.8440       & 
                                            4.6252        & 
                                            6.0166         & 
                                            5.3825  \\
                               & Naive WL & \textbf{\qquad +12.9973} & 
                                \qquad \textbf{+6.0987}  & 
                                \qquad \textbf{\textit{+0.6665}} &
                                \qquad \textbf{\textit{+0.2998}} &
                                \qquad \textbf{+0.0226}  \\
                               & C-WL     & \textbf{\qquad \textit{+13.2291}}& 
                                \qquad \textbf{\textit{+6.1149}} & 
                                \qquad \textbf{+0.4436}  & 
                                \qquad -0.2615      & 
                                \textbf{\qquad  \textit{+0.1322}} \\
                               & G-WL     & \textbf{\qquad +11.6967} &  
                                N. A.   &
                                \qquad \textbf{+0.4553}  & 
                                \qquad -0.8769         & 
                                \qquad \textbf{+0.1061}  \\
        \hline
        \multirow{4}{*}{QM8}   & Atomic & 0.0310 & 
                                          0.0234 & 
                                          0.0209 & 
                                            0.0322         & 
                                            0.0365  \\
                               & Naive WL & \textbf{\qquad +0.0006} & 
                                \textbf{\qquad \textit{+0.0030}} & 
                                \qquad -0.0029          & 
                                \textbf{\qquad +0.0005} & 
                                \textbf{\qquad \textit{+0.0032}} \\
                               & C-WL     & \textbf{\qquad +0.0004} & 
                                \textbf{\qquad +0.0020} & 
                                \textbf{\qquad \textit{+0.0009}}& 
                                \qquad 0.0000  & 
                                \textbf{\qquad +0.0025}  \\
                               & G-WL     & \textbf{\qquad \textit{+0.0018}}& 
                                \textbf{\qquad +0.0029} & 
                                \textbf{\qquad +0.0004} & 
                                \textbf{\qquad \textit{+0.0006}} & 
                                \textbf{\qquad +0.0027} 
    \end{tabular}
    \caption{Ten-run averages of mean absolute errors (MAEs) of the regression benchmarks. For the atomic embedding, we present the test MAEs. 
    For WL embeddings, we present the error reduction from the atomic embedding i.e., positive values indicate the performance improvements. 
    For example, the MAE of GCN+atomic embeedding on LIPO dataset (at the upper-left corner) is 0.7077. That of GCN+naive WL is \textit{better (smaller)}: $0.6333=0.7077-0.0744$. 
    We highlight the results where the performance was improved in bold and the best results of each dataset in italics. For the QM9 dataset, we couldn't obtain the result of RelGAT + G-WL due to the large data size and intensive memory usage. The full table with standard deviations can be found in Supplemental material. }
    \label{tab:optimized_result_regression}
\end{table*}

\begin{table*}[t]
    \centering
    \small
    \begin{tabular}{cc||c@{\hskip 0.5\tabcolsep}c@{\hskip 0.5\tabcolsep}c@{\hskip 0.5\tabcolsep}c@{\hskip 0.5\tabcolsep}c}    
         Dataset & Embedding & GCN &
                               RelGAT &
                               GGNN &
                               GIN &
                               NFP \\
         \hline 
         \multirow{4}{*}{Tox21} & Atomic & 0.7082 & 
                                            0.7432 & 
                                            0.7136 & 
                                            0.7237   &  
                                            \textit{0.7618} \\
                                & Naive WL & \textbf{\qquad +0.0606} & 
                                                \qquad -0.0499           & 
                                                \textbf{\qquad +0.0142}  & 
                                                \qquad -0.0251         & 
                                                \qquad -0.0425   \\
                                & C-WL     & \textbf{\qquad +0.0417}  & 
                                                \textbf{\qquad \textit{+0.0177}} & 
                                                \textbf{\qquad \textit{+0.0548}} & 
                                                \textbf{\qquad +0.0160} & 
                                                \qquad -0.0193 \\
                                & G-WL     & \textbf{\qquad \textit{+0.0625}} & 
                                                \qquad -0.0013  & 
                                                \textbf{\qquad +0.0442}  & 
                                                \textbf{\qquad \textit{+0.0347}} & 
                                                \qquad -0.0312   \\
        \hline
        \multirow{4}{*}{HIV} & Atomic & 0.7507 & 
                                        0.7037 & 
                                        0.7044 & 
                                        0.7108         & 
                                        0.6533   \\
                             & Naive WL &   \qquad -0.0221 & 
                                            \qquad -0.0098 & 
                                            \textbf{\qquad \textit{+0.0142}} & 
                                            \qquad -0.0064         & 
                                            \textbf{\qquad \textit{+0.0988}}  \\
                             & C-WL     &   \textbf{\qquad\textit{+0.0038}} & 
                                            \textbf{\qquad \textit{+0.0206}} & 
                                            \textbf{\qquad +0.0139} & 
                                            \textbf{\qquad \textit{+0.0134}}  & 
                                            \textbf{\qquad +0.0593}  \\
                             & G-WL     &   \qquad -0.0127 & 
                                            \qquad \textbf{+0.0080} & 
                                            \qquad \textbf{+0.0087} & 
                                            \qquad -0.0070         & 
                                            \qquad \textbf{+0.0984}  \\
        \hline
        \multirow{4}{*}{ToxCast} & Atomic & 0.7745 & 
                                            0.7820 &
                                            0.7720 & 
                                            0.7691         & 
                                            0.7585   \\
                                 & Naive WL &   \textbf{\qquad \textit{+0.0182}}& 
                                                \qquad -0.0358          & 
                                                \qquad -0.0222          & 
                                                \textbf{\qquad +0.0043} & 
                                                \qquad -0.0005   \\
                                 & C-WL     &   \textbf{\qquad +0.0086} & 
                                                \textbf{\qquad \textit{+0.0193}}& 
                                                \textbf{\qquad +0.0034} & 
                                                \textbf{\qquad \textit{+0.0103}} & 
                                                \textbf{\qquad \textit{+0.0123}} \\
                                 & G-WL     &   \textbf{\qquad +0.0043} & 
                                                \qquad -0.0017 & 
                                                \textbf{\qquad \textit{+0.0155}} & 
                                                \textbf{\qquad +0.0053}  & 
                                                \qquad -0.0284       \\
     \hline
        \multirow{4}{*}{clintox} & Atomic & 0.9387 & 
                                            0.9287 & 
                                            0.9338 & 
                                            0.9305 & 
                                            0.9214  \\
                                 & Naive WL &   \qquad -0.0017 & 
                                                \qquad \textbf{+0.0071} & 
                                                \qquad \textbf{+0.0091} & 
                                                \qquad -0.0050 & 
                                                \qquad \textbf{+0.0122} \\
                                 & C-WL &       \qquad \textbf{\textit{+0.0158}} & 
                                                \qquad \textbf{\textit{+0.0072}} & 
                                                \qquad \textbf{+0.0100} & 
                                                \qquad \textbf{\textit{+0.0026}} & 
                                                \qquad \textbf{\textit{+0.0195}} \\
                                 & G-WL &       \qquad -0.0034 & 
                                                \qquad -0.0082 & 
                                                \textbf{\qquad \textit{+0.0173}} & 
                                                \qquad -0.0014 & 
                                                \textbf{\qquad +0.0144} \\
        \hline
        \multirow{4}{*}{muv} & Atomic & 0.6350 & 
                                        0.6164 & 
                                        0.6586 & 
                                        0.5716 &
                                        0.6839  \\
                                 & Naive WL &   \textbf{\qquad \textit{+0.0778}}& 
                                                \textbf{\qquad \textit{+0.0560}} & 
                                                \textbf{\qquad +0.0738} & 
                                                \textbf{\qquad \textit{+0.0786}} & 
                                                \textbf{\qquad \textit{+0.0523}} \\
                                 & C-WL &       \textbf{\qquad +0.0503} & 
                                                \textbf{\qquad +0.0513} &  
                                                \textbf{\qquad \textit{+0.0931}} & 
                                                \textbf{\qquad +0.0055} &  
                                                \textbf{\qquad +0.0288}  \\
                                 & G-WL &       \textbf{\qquad +0.0456} & 
                                                \textbf{\qquad +0.0223} & 
                                                \textbf{\qquad +0.0326} & 
                                                \textbf{\qquad +0.0287} &  
                                                \textbf{\qquad +0.0278} 
          \end{tabular}   
    \caption{Ten-run averages of ROC-AUCs for the classification benchmarks. The table format is the same as Table~\ref{tab:optimized_result_regression}, excepting that AUC gains from the atomic embedding are presented for WL embeddings. 
    For example, the ROC-AUC of GCN+atomic embeedding on Tox21 dataset (at the upper-left corner) is 0.7082. That of GCN+naive WL is \textit{better (larger)}: $0.7688=0.7082+0.0606$. }
    \label{tab:optimized_result_classification}
\end{table*}

\section{Conclusion}

In this study, we considered the design of molecular features for GNNs.
Based on the history of chemical studies, we proposed the use of WL embedding, a simple embedding method leveraging local atomic patterns. 
We also proposed its variants, C-WL and G-WL embeddings, which alleviate the sparsity problem that might arise in the naive WL embedding. 
We showed that the representation power of WL embedding is equivalent to that of atomic embedding + ReLU GNN layer while reducing the parameter scale. 
The efficacy of the WL embeddings, especially C-WL embedding, was presented in intensive experimental validations against multiple GNN architectures and multiple benchmark molecular graph datasets with different chemical properties.

\paragraph{Acknowledgments.}
We would like to thank Kenji Fukumizu for insightful discussions.


\bibliography{neurips_2020}
\bibliographystyle{plainnat}

\appendix

\newpage

\begin{center}
    \LARGE Supplementary Material for\\ ``Weisfeiler-Lehman Embedding for Molecular Graph Neural Networks''
\end{center}
\vspace{3em}
\section{Additional Related Work}

\subsection{Improving GNNs for Multiple Datasets and Architectures}

After the seminal study of GCN~\citep{Kipf_Welling17ICLR}, many GNN architectures have been proposed, which introduce inductive biases suit to a dataset of interest. 
Unfortunately, the community still is not aware of the golden-standard GNN architecture that works fine across datasets and downstream tasks~\citep{Wu18MoleculeNet,Ishiguro19arXiv}. 
This makes a clear contrast with ResNet~\citep{He16CVPR} in the computer vision community. Therefore users of GNNs must choose the good GNN architecture for each dataset and downstream task. 

A few researchers are interested in architecture-free techniques that improve performances of existing GNNs across several datasets and types of downstream tasks. 
\citet{Hu20ICLR} proposed a strategy for pre-training GNNs with massive unsupervised and smaller supervised datasets. In the experiments, pre-trained GNNs obtained better performances over chemical and biological graph datasets. 
\citet{Ishiguro19arXiv} developed a sub-network module that can be attached to generic MPNNs. The module bypasses neighborhood-based information propagation via a virtual super-node and gate mechanisms. The module consistently improves the performances of six different GNNs for molecular graph datasets.  

We share the same goal with these works. We design the embedding so as not to fine-tune to specific a GNN architecture. Our experiments demonstrate that the proposed embedding consistently improves the performances of different GNNs for multiple molecular graph datasets. 

\subsection{GNNs inspired by WL algorithm}

\citet{Jin17NIPS,Lei17ICML} propose MPNNs whose layer update rules are inspired by mimicking the WL kernel. The main idea is to concatenate the latent vectors of the neighboring nodes. 

\citet{Xu19ICLR,Morris19AAAI} reveal that the WL algorithm has the upper bound capability of the MPNNs, in terms of isomorphism testing. 
\citet{Xu19ICLR} proves that the bijectivity is required for layer update functions to achieve the upper bound. The authors propose a GIN architecture where the function in each layer is modeled by Multi-layer Perceptrons. \citet{Morris19AAAI} proposes to emulate the $k (>1)$-dimensional WL algorithm for further capability. 
As explained in the main manuscript, all of these previous works propose new GNN architecture (layer formulations) but do not attend the raw node feature representations \textit{before} applying GNN layers. 

\subsection{Rich Atom Information}

If the molecular graphs of interest provide additional features (e.g., electronic charge of an atom) other than discrete node labels, we can leverage these additional features. 
Some GNN studies propose to combine such features with their network architectures (e.g.,~\citep{Yang19arXiv,Pham17arXiv}). 
Recently, a few GNN studies are interested in inferring the (imperfect) 3D structure of molecule graphs based on distance information between atom nodes~\citep{Takamoto19arXiv,Klicpera20ICLR}. 
These approaches are orthogonal to our proposal. We can augment these GNN studies by simply concatenating the node label embedding by our proposed method as an additional feature for nodes. 

\subsection{Other topics}

\citet{Liu19NeurIPS} proposed an interesting and unique approach for molecular graph analysis, based on the Bag of Random-walks approach. 
This method computes the graph representation based on the Bag of Random-walk embeddings. 
The embedding is based on the node representation. 
This method first trains the NN to compute the node embedding in an unsupervised manner. The training objective is based on the loss of CBoW~\citep{Mikolov13} in NLP tasks. Namely, a node's embedding vector is determined to be predictable from the neighboring nodes' embedding vectors.

\section{1-Dimensional Weisfeiler--Lehman Algorithm}

The $k$-Dimensional WL algorithm~\citep{Weisfeiler_Lehman68} is a heuristic used to identify a (non-)isomorphism between two graphs, $G_1$ and $G_2$, based on $k$-tuples in the graphs.

The most popular algorithm is the $1$-dimensional WL algorithm. This algorithm expands a label of each node by concatenating the labels of neighboring nodes and relabels (hash) them in a new label. 
The algorithm consists of three steps~\citep{Shervashidze11,Morris19AAAI}: 
\begin{enumerate}
    \item \textbf{Collecting neighbor labels:} For each node $v_{i}$, labels in $\mathcal{N}_{i}$ are collected and form a multiset of labels, $\mathcal{M}_{i}$. 
    \item \textbf{Expanding the label:} $\mathcal{M}_i$ is attached to $\ell_{i}$ to obtain the expanded label $s_{i}$. Namely, $s_{i} \leftarrow (\ell_{i}, \mathcal{M}_{i})$.  
    \item \textbf{Relabeling:} $\ell_{i}$ is relabeled by hashing $s_i$:
    $\ell_{i} \leftarrow \text{HASH} (s_{i})$, where HASH($\cdot$) bijectively maps $s_{i}$ to a unique value in a label set $\Sigma$. If $s_{i}$ first appears within the computation, the HASH augments $\Sigma$ with a new symbol that is not seen in $\Sigma$ and returns that new symbol. 
\end{enumerate}
Performing the above steps for all nodes in two graphs, the WL algorithm tests the following termination condition: If there is a node label that only appears in one graph, then the two graphs are not isomorphic. Otherwise, the WL algorithm repeats the above steps. 

\section{Proof of Theorem~\ref{thm:relu}}
Our goal is to construct a one-to-one mapping from lattice points to linearly independent vectors. We first fix $k$ in \eqref{eq:relu_gnn} to an arbitrary number so that we can focus on the single lattice $L(\mathbf{m})$, and then we generalize the result for all $k=1,\dots,K$. Let $H$ be a $(M+1)^K \times d_1$ matrix whose row contains $h_i$ of each lattice point. Our strategy is to make $H$ triangular with non-zero diagonal elements so that all the row vectors are automatically linearly independent with sufficiently large $d_1$. 

Suppose we employ the $K$-dimensional standard basis scaled by $\alpha>0$ as the embedding vectors, i.e., $(x_{k})_l=\alpha\delta_{kl}$. Now $b$ in \eqref{eq:relu_gnn} and $E = (e_{1},\dots,e_K)$, the basis of $L$, are directly determined by $U, V$, and $\alpha$,  and without loss of generality we consider the parameterization $\{W, \alpha b, \alpha E\}$ instead of $\{W, U, V, X\}$. 
Let $E$ be again the $K$-dimensional standard basis, and we simply have $L(\mathbf{m})=\alpha \mathbf{m}$ for $\mathbf{m}\in\{0,1,\dots,M\}^K$. As a weight vector, we consider $w = ((M+1)^{K-1}, (M+1)^{K-2},\dots, M+1,1)$ so that the inner product $\langle w, x\rangle$ calculates the number of $\mathbf{m}$ in base $M+1$ (e.g., $\langle w, \mathbf{m}\rangle = 5$ for $\mathbf{m}=(0,1,0,1)$ when $M=1$, which is equivalent to the number of the binary bits $0101$). By using this idea, we can set the weight $W$ and the bias $b$ as 
\begin{align}
    W = \begin{pmatrix}
    (M+1)^{K-1} & \cdots & M+1 & 1\\
    (M+1)^{K-1} & \cdots & M+1 & 1\\
    \vdots &\vdots &\ddots &\vdots\\
    (M+1)^{K-1} & \cdots & M+1 & 1
    \end{pmatrix},
    \qquad
    b =\begin{pmatrix}
    (M+1)^{K} - 1\\
    (M+1)^{K} - 2\\
    \vdots\\
    0\\
    -1
    \end{pmatrix}.
\end{align}
We see that the number of nonzero elements of $h=\sigma(WL(\mathbf{m}) - \alpha b)=\alpha\sigma(W\mathbf{m} - b)$ corresponds to the number of $\mathbf{m}$ in base $M+1$ and $H$ becomes upper triangular, i.e.,
\begin{align}
    H = \alpha\sigma\left(
    \mathbf{M} W^\top-
    \begin{pmatrix}
    b^\top\\
    \vdots\\
    b^\top
    \end{pmatrix}
    \right)
    = \alpha\begin{pmatrix}
    1 & \cdots &(M+1)^{K-1} & \cdots & (M+1)^{K} -1 & (M+1)^K\\
    \vdots&\ddots&\vdots&\ddots&\vdots&\vdots\\
    0 & \cdots &1 & \cdots & M & M+1\\
    0 & \cdots &0 & \cdots & M-1 & M\\
    \vdots&\ddots&\vdots&\ddots&\vdots&\vdots\\
    0 & \cdots &0 & \cdots & 1 & 2\\
    0 & \cdots &0 & \cdots & 0 & 1\\
    \end{pmatrix}
\end{align}
where $\mathbf{M}$ is a matrix whose $i$-th from the bottom contains the $K$ digits of $i$ in base $M+1$, i.e.,
\begin{align}
    \mathbf{M} = 
    \begin{pmatrix}
    M & \dots &M & M & M\\
    \vdots&\ddots&\vdots&\vdots&\vdots\\
    0 & \dots &0 & 1 & 0\\
    0 & \dots &0 & 0 & M\\
    \vdots&\ddots&\vdots&\vdots&\vdots\\
    0 & \dots &0 & 0 & 1\\
    0 & \dots &0 & 0 & 0\\
    \end{pmatrix}.
\end{align}
Hence such $\{W, \alpha b, \alpha E\}$ meet our goal.

Next, we measure their scale. The $\ell_2$-norm of $b$ is given by the sum of the squares of the sequence from $-1$ to $(M+1)^K - 1$. Substituting $n=M^K$ into the formula $\sum_{i=1}^n i^2 = \frac{n (n+1)(2n+1)}{6}$ and since $\alpha$ is constant, we conclude that $\|\alpha b\|_2=\alpha\| b\|_2=\Omega(M^{3K/2})$. The same thing holds for $\alpha \mathbf{M} W^\top$.

To generalize this for all $k=1,\dots,K$, we can extend $W$ and $E$ as block-diagonal where each block of them contain their original contents, which keep $H$ linearly independent for all lattices.

\section{Dataset Specification}

\subsection{MoleculeNet Datasets}

MoleculeNet~\citep{Wu18MoleculeNet} is a standard collection of molecular graph datasets. It contains several molecular datasets from four different chemical fields: quantum chemistry, physical chemistry, physiology, and biophysics. 

\subsubsection{Dataset Details}
For the graph regression tasks, we used the QM9 dataset, the QM8 dataset, and the Lipophilicity (LIPO) dataset.
QM9 is a dataset containing approximately 133K molecules with 9 types of heavy atoms. The dataset consists of 12 important numerical values (target variables) of the chemical-energetic, electronic, and thermodynamic properties, such as HOMO, LUMO, and electron gaps computed through quantum chemistry techniques.
QM8 is another quantum chemical dataset containing approximately 22K molecules with eight types of heavy atoms. QM8 also consists of 12 important chemical-energetic, electronic, and thermodynamic properties; however, different quantum chemistry methods are used for the computations. 
The LIPO dataset contains the solubility values of roughly 4K drug molecules. This dataset is taken from the physical chemistry field. 
Each sample in these datasets is a pair of a molecular graph and a numerical value(s): the 12 chemical properties in the QM9 and QM8 dataset, and the solubility in the LIPO dataset.
For these datasets, the task is to predict the numerical value(s) from the molecular graph. 
We evaluated the performance of the models using mean absolute errors (MAEs). 
We report the averaged MAE over 12 sub-tasks (properties) for QM9 and QM8. 

For the graph classification tasks, we used the Tox21, the HIV, the ToxCast, the Clintox, and the MUV datasets. 
The Tox21 dataset contains approximately 8K pairs of a molecular graph and a 12-dimensional binary vector that represents the experimental outcomes of 
the toxicity measurements on 12 different targets. 
The HIV dataset contains roughly 42K pairs of a molecular graph and a binary label that represent the medicinal effect of the molecule. 
The ToxCast dataset contains about 8K pairs of a molecular graph and a 617-dimensional binary vector that represent the different experimental results. 
The Clintox dataset contains 1491 pairs of a molecular graph and two binary labels that represent the clinical toxicity and the FDA approval status. 
The MUV (Maximum Unbiased Validation) dataset contains 93 molecular graph samples attached with 17 labels, specifically designed for virtual screening validations. 
For these datasets, the task is to predict the binary label(s) from the molecular graph. 
The Tox21, ToxCast, and the Clintox datasets are taken from the chemical physiology field. HIV and the MUV datasets belong to the biophysics collection of MoleculeNet. 
For these tasks, we use ROC-AUC values as a measure of performance. 
We report the averaged ROC-AUC over 12 sub-tasks (targets) for Tox21, 617 sub-tasks for ToxCast, 2 sub-tasks for Clintox, and 17 sub-tasks for MUV, respectively. 

\subsubsection{Graph Data Representation}

All real-world datasets used in our experiments are provided in the SMILES format. 
A SMILES format is a line notation for describing the structure of chemical compounds. 
We decode a SMILES molecular data into a multigraph representation of the molecule. 
A node in the graph corresponds to an atom. Each atom node is associated with the symbolic label of the atom name (``H'', ``C'', ...). 

An edge in the graph corresponds to a bond between atoms. Each bond edge is associated with the bond type information (single, double, ....). The edges $\mathcal{E}$ have connectivity information between nodes (topology), as well the bond type information. 

\subsubsection{Data splits}
MoleculeNet provides several ways of data splitting. 
The ``random'' split is the random sample shuffling that is most familiar to the machine learning community. 
The ``scaffold'' split separate samples based on the molecular two-dimensional structure. Since the scaffold split separates structurally different molecules into different subsets, ``it offers a greater challenge for learning algorithms than the random split''~\citep{Wu18MoleculeNet}. 
Throughout the paper, we adopt the scaffold split to assess the full potential of the GWM-attaching GNNs. 

The actual construction of the scaffold split train/validation/test subsets has freedom of algorithm choices. 
We adopted the algorithm provided by the deepchem\footnote{https://deepchem.io/} library, which is the standard split algorithm for many papers. 

\section{Model Specification}

\subsection{Readout Layer}

In many applications of GNNs users may expect a single fixed-length vector representing the characteristics of the graph $G$. 
So we add the 'readout' layer to aggregate the node latent vectors. 

The main issue in the readout unit is how to aggregate nodes, whose number varies for each graph. 
A simple way is to take an arithmetic average (sum) of the node representations at the $L$-th layer, but we can also use a DNN to compute (non-linear) ``average'' of the node representations~\citep{Li16ICLR,Gilmer17ICML}.  
In the experiments, we implement the readout layer that is described in the original papers of each GNN architecture. 

\subsection{Hyperparameters}

In the main experiments, we optimize three hyperparameters. (i) $L$: the number of layers of GNNs, (ii) $D$: the dimension of latent hidden vectors of nodes in GNN layers, and (iii) $\alpha$: the step size of Adam~\citep{Kingma_Ba15ICLR} optimizer. 

We optimize these hyperparameters using Optuna~\citep{Akiba19SIGKDD} library. Target functions to be optimized is MAE (for regression) or ROC-AUC (for classification) on the validation dataset. 
Tables~\ref{tab:optimized_hyprms_regression_dc},\ref{tab:optimized_hyprms_classification_dc} present the chosen $L$, $D$, and $\alpha$. 
The maximum $D$ of RelGAT is set to 32, to avoid the memory shortage errors. 

Other hyperparameters are fixed and shared among experiemnts. 
We trained each GNNs for 500 epochs. The minibatch size is fixed for 128 for GCN and GGNN, and 32 for RelGAT because of the GPU memory limits. 
All models were trained with Adam~\citep{Kingma_Ba15ICLR}, $\beta_{1} = 0.9$, and $\beta_{2}=0.999$.

\begin{table*}[t]
    \centering
    \footnotesize
    \begin{tabular}{cc||ccccc}
         Dataset & Embedding & GCN & RelGAT & GGNN & GIN & NFP\\ \hline 
         \multirow{4}{*}{LIPO} & Atomic & (110, 3, 0.0017) & (16, 4, 0.0022) & (37, 3, 0.0091) & (81, 3, 0.0015) & (53, 5, 0.0062) \\
         & Naive WL & (124, 2, 0.0018) & (21, 3, 0.0038) & (52, 2, 0.013) & (37, 2, 0.0064) & (176, 5, 0.00095) \\
         & C-WL & (188, 2, 0.00016) & (23, 4, 0.0026) &  (45, 5, 0.0032) & (17, 2, 0.0016) & (233, 2, 0.00017) \\
         & G-WL & (164, 2, 0.0015) & (21, 4, 0.0034) & (45, 4, 0.0080) & (39, 3, 0.0014) & (232, 5, 0.00041) \\
                                 \hline
         \multirow{4}{*}{QM9} & Atomic & (140, 2, 0.00031) & (41, 4, 0.0022) & (232, 5, 0.00029) & (256, 4, 0.00016) & (135, 6, 0.00049) \\
         & Naive WL & (234, 2, 0.00047) & (205, 3, 0.00069) & (254, 6, 0.00018) & (214, 2, 0.00032) & (73, 5, 0.00081) \\
         & C-WL & (253, 2, 0.00020) & (56, 3, 0.0021) &  (248, 6, 0.00014) & (255, 2, 0.00014) & (162, 5, 0.00035) \\
         & G-WL & (229, 3, 0.00011) & N.A. &  (237, 6, 0.00035) & (167, 2, 0.00055) & (256, 6, 0.00034) \\
                                          \hline
         \multirow{4}{*}{QM8} & Atomic & (148, 5, 0.00096) & (32, 4, 0.0014) & (122, 5, 0.00097) & (253, 2, 0.00023) & (253, 5, 0.00033) \\
         & Naive WL & (93, 4, 0.0021) & (88, 6, 0.00056) & (174, 4, 0.0013) & (137, 2, 0.00028) & (139, 5, 0.00015) \\
         & C-WL & (137, 5, 0.00011) & (58, 5, 0.00097) & (150, 3, 0.0010) & (300, 2, 0..00023) & (162, 6, 0.00024) \\
         & G-WL & (254, 5, 0.00023) & (16, 3, 0.0018) & (277, 2, 0.0013) & (86, 3, 0.00018) & (255, 5, 0.000049)
    \end{tabular}
    \caption{Hyperparameter choices for the DeepChem-scaffold splits, regression tasks. 
    Each cell is formatted as ($D$, $L$, $\alpha$) where $D$ denotes the dimension of latent vectors, $L$ denotes the number of layers,  and $\alpha$ denotes the step size of Adam. }
    \label{tab:optimized_hyprms_regression_dc}
\end{table*}

\begin{table*}[t]
    \centering
    \footnotesize
    \begin{tabular}{cc||ccccc}
         Dataset & Embedding & GCN & RelGAT & GGNN & GIN & NFP \\ \hline 
         \multirow{4}{*}{Tox21} & Atomic & (22, 4, 0.00024) & (15, 3, 0.0027) & (32, 3, 0.0018) & (126, 4, 0.0012) & (66, 2, 0.0011) \\
         & Naive WL & (112, 2, 0.000088) & (22, 4, 0.00042) & (52, 2, 0.0029) & (29, 2, 0.012) & (119, 2, 0.0057) \\
         & C-WL & (235, 3, 0.0023) & (14, 2, 0.000054) & (27, 3, 0.0055) & (49, 5, 0.0027) & (80, 3, 0.0039) \\
         & G-WL & (51, 4, 0.0020) & (7, 2, 0.0084) & (75, 4, 0.0090) & (55, 4, 0.00036) & (78, 4, 0.0091) \\
                                 \hline
         \multirow{4}{*}{HIV} & Atomic & (105, 4, 0.00022) & (6, 2,  0.0097) & (23, 2, 0.0025) & (74, 2, 0.00021) & (56, 4, 0.0031) \\
         & Naive WL & (50, 5, 0.0030) & (16, 4, 0.0020) & (27, 4, 0.000042) & (28, 2, 0.00041) & (230, 3, 0.000016) \\
         & C-WL & (86, 6, 0.0011) & (20, 3, 0.000065) &  (45, 2, 0.0017) & (19, 2, 0.0031) & (57, 3, 0.0020 \\
         & G-WL & (43, 4, 0.0016) & (5, 3, 0.000053) & (110, 3, 0.0027) & (42, 2, 0.0018) & (39, 2, 0.00020) \\
           \hline
         \multirow{4}{*}{ToxCast} & Atomic & (177, 3, 0.000017) & (20, 3, 0.0068) & (30, 4, 0.0021) & (80, 3, 0.0024) & (121, 2, 0.0020) \\
         & Naive WL & (73, 2, 0.00020) & (17, 3, 0.00079) & (71, 5, 0.0028) & (25, 3, 0.019) & (25, 2, 0.0022) \\
         & C-WL & (103, 5, 0.00039) & (20, 3, 0.0050) & (128, 4, 0.0046) & (64, 3, 0.0052) & (79, 2, 0.00035) \\
         & G-WL & (80, 2, 0.0037) & (15, 3, 0.0110) & 47, 2, 0.0153) & (81, 4, 0.00016) & (185, 2, 0.0020) \\
           \hline
         \multirow{4}{*}{clintox} & Atomic & (135, 5, 0.0034) & (4, 3, 0.014) & (92, 5, 0.0040) & (34, 3, 0.0046) & (17, 4, 0.0094) \\
         & Naive WL & (99, 4, 0.00049) & (11, 3, 0.078) & (19, 6, 0.0031) & (38, 5, 0.0029) & (49, 3, 0.032) \\
         & C-WL & (74, 3, 0.00063) & (5, 3, 0.036) & (24, 4, 0.0060) & (124, 3, 0.0015) & (131, 2, 0.0021) \\
         & G-WL & (71, 4, 0.021) & (6, 2, 0.020) & (75, 5, 0.00076) & (102, 6, 0.010) & (135, 5, 0.021) \\
           \hline
         \multirow{4}{*}{muv} & Atomic & (36, 3, 0.00021) & (8, 4, 0.0012) & (50, 5, 0.00041) & (214, 4, 0.000021) & (28, 5, 0.0017) \\
         & Naive WL & (55, 3, 0.00064) & (11, 3, 0.0018) & (148, 2, 0.00013) & (104, 2, 0.000086) & (178, 3, 0.000065) \\
         & C-WL & (102, 3, 0.00011) & (22, 4, 0.00053) & (154, 5, 0.000016) & (101, 3, 0.000018) & (55, 5, 0.00034) \\
         & G-WL & (143, 3, 0.000076) & (22, 3, 0.0012) & (24, 3, 0.0012) & (20, 4, 0.000015) & (169 5, 0,00019)
    \end{tabular}
    \caption{Hyperparameter choices for the DeepChem-scaffold splits, classification tasks. 
    Each cell is formatted as ($D$, $L$, $\alpha$) where $D$ denotes the dimension of latent vectors, $L$ denotes the number of layers,  and $\alpha$ denotes the step size of Adam. }
    \label{tab:optimized_hyprms_classification_dc}
\end{table*}

\section{Full Results on Molecular Graph Experiments}

Tables~\ref{tab:optimized_result_regression_full},\ref{tab:optimized_result_classification_full} present the full results of the Tables~\ref{tab:optimized_result_regression},\ref{tab:optimized_result_classification} in the molecular graph data experiments. 

\begin{table*}[t]
    \centering
    \small
    \begin{tabular}{c@{\hskip 0.9\tabcolsep}c||r@{\hskip 0.05\tabcolsep}c@{\hskip 0.9\tabcolsep}r@{\hskip 0.05\tabcolsep}c@{\hskip 0.9\tabcolsep}r@{\hskip 0.05\tabcolsep}c@{\hskip 0.9\tabcolsep}r@{\hskip 0.05\tabcolsep}c@{\hskip 0.9\tabcolsep}r@{\hskip 0.05\tabcolsep}c}
         Dataset & Embedding & \multicolumn{2}{c}{GCN} &
                               \multicolumn{2}{c}{RelGAT} &
                               \multicolumn{2}{c}{GGNN} &
                               \multicolumn{2}{c}{GIN} &
                               \multicolumn{2}{c}{NFP} \\
        \hline 
        \multirow{4}{*}{LIPO} & Atomic    & 0.7077         & \tiny{$\pm$0.0154} &
                                            0.6298         & \tiny{$\pm$0.0208} &
                                            0.6670         & \tiny{$\pm$0.0206} &
                                            0.7034         & \tiny{$\pm$0.0251} & 
                                            0.7067         & \tiny{$\pm$0.0883} \\
                               & Naive WL & \textbf{\textit{0.6333}} & \tiny{$\pm$0.0112} &
                                            \textbf{0.5836}  & \tiny{$\pm$0.0311} &
                                            0.7386           & \tiny{$\pm$0.0960} &
                                            \textbf{\textit {0.6393}} & \tiny{$\pm$0.0164} & 
                                            \textbf{0.6485} & \tiny{$\pm$0.0194} \\
                               & C-WL     & \textbf{0.6524}  & \tiny{$\pm$0.0165} &
                                            \textbf{\textit{0.5811}} & \tiny{$\pm$0.0143} &
                                            \textbf{\textit{0.5963}} & \tiny{$\pm$0.0251} &
                                            \textbf{0.6855} & \tiny{$\pm$0.0228} & 
                                            \textbf{\textit{0.6354}}         & \tiny{$\pm$0.0172} \\
                               & G-WL     & \textbf{0.6389}  & \tiny{$\pm$0.0198} &   
                                            \textbf{0.5845}  & \tiny{$\pm$0.0169} &
                                            \textbf{0.6415}  & \tiny{$\pm$0.0229} &
                                            \textbf{0.6842} & \tiny{$\pm$0.0481} & 
                                            \textbf{0.6485} & \tiny{$\pm$0.0105} \\
        \hline
        \multirow{4}{*}{QM9}   & Atomic   & 23.7247       & \tiny{$\pm$1.2825} &
                                            11.8440       & \tiny{$\pm$0.7028} &
                                            4.6252        & \tiny{$\pm$0.5631} &
                                            6.0166         & \tiny{$\pm$0.2230} & 
                                            5.3825        & \tiny{$\pm$0.1375} \\
                               & Naive WL & \textbf{10.7274} & \tiny{$\pm$0.6118} & 
                                            \textbf{5.7453}  & \tiny{$\pm$0.1262} & 
                                            \textbf{\textit{3.9587}}  & \tiny{$\pm$0.0862} &
                                            \textbf{\textit{5.7168}} & \tiny{$\pm$0.0919} & 
                                            \textbf{5.3599}  & \tiny{$\pm$0.2018} \\
                               & C-WL     & \textbf{\textit{10.4956}}& \tiny{$\pm$0.6118} & 
                                            \textbf{\textit{5.7291}} & \tiny{$\pm$0.3087} & 
                                            \textbf{4.1816}  & \tiny{$\pm$0.1758}  &
                                            6.2781         & \tiny{$\pm$0.3279} & 
                                            \textbf{\textit{5.2503}} & \tiny{$\pm$0.1602} \\
                               & G-WL     & \textbf{12.0280} & \tiny{$\pm$1.7032} & 
                                                            \multicolumn{2}{c}{N. A.}   &
                                            \textbf{4.1699}  & \tiny{$\pm$0.1256} &
                                            6.8935         & \tiny{$\pm$0.2965} & 
                                            \textbf{5.2764}  & \tiny{$\pm$0.1311} \\
        \hline
        \multirow{4}{*}{QM8}   & Atomic & 0.0310 & \tiny{$\pm$0.0004} &
                                          0.0234 & \tiny{$\pm$0.0005} &
                                          0.0209 & \tiny{$\pm$0.0002} &
                                            0.0322         & \tiny{$\pm$0.0011} & 
                                            0.0365         & \tiny{$\pm$0.013} \\
                               & Naive WL & \textbf{0.0304} & \tiny{$\pm$0.0009} &
                                            \textbf{\textit{0.0204}}& \tiny{$\pm$0.0004} &
                                            0.0238          & \tiny{$\pm$0.0012} &
                                            \textbf{0.0317} & \tiny{$\pm$0.0009} & 
                                            \textbf{\textit{0.0333}} & \tiny{$\pm$0.0082} \\
                               & C-WL     & \textbf{0.0306} & \tiny{$\pm$0.0002} & 
                                            \textbf{0.0214} & \tiny{$\pm$0.0014} & 
                                            \textbf{\textit{0.0200}}& \tiny{$\pm$0.0002}  &
                                            0.0322         & \tiny{$\pm$0.0004} & 
                                            \textbf{0.0340} & \tiny{$\pm$0.0007} \\
                               & G-WL     & \textbf{\textit{0.0292}}& \tiny{$\pm$0.0002} & 
                                            \textbf{0.0205} & \tiny{$\pm$0.0004} & 
                                            \textbf{0.0205} & \tiny{$\pm$0.0004} &
                                            \textbf{\textit{0.0316}} & \tiny{$\pm$0.0007} & 
                                            \textbf{0.0338} & \tiny{$\pm$0.0013} 
    \end{tabular}
    \caption{Ten-run averages of mean absolute errors for the regression benchmarks. Smaller values are better. Note that the target values are not normalized across attributes. Bold indicates that WL embedding improves the generalization performance over atomic embedding. Italics indicate the best embedding within each cell. We couldn't obtain the result of RelGAT + G-WL for the QM9 dataset (``N.A.'') due to the large data size and the memory-intensive formulation of the RelGAT model. }
    \label{tab:optimized_result_regression_full}
\end{table*}

\begin{table*}[t]
    \centering
    \small
    \begin{tabular}{c@{\hskip 0.9\tabcolsep}c||r@{\hskip 0.05\tabcolsep}c@{\hskip 0.9\tabcolsep}r@{\hskip 0.05\tabcolsep}c@{\hskip 0.9\tabcolsep}r@{\hskip 0.05\tabcolsep}c@{\hskip 0.9\tabcolsep}r@{\hskip 0.05\tabcolsep}c@{\hskip 0.9\tabcolsep}r@{\hskip 0.05\tabcolsep}c}    
         Dataset & Embedding & \multicolumn{2}{c}{GCN} &
                               \multicolumn{2}{c}{RelGAT} &
                               \multicolumn{2}{c}{GGNN} &
                               \multicolumn{2}{c}{GIN} &
                               \multicolumn{2}{c}{NFP} \\
         \hline 
         \multirow{4}{*}{Tox21} & Atomic & 0.7082 & \tiny{$\pm$0.0098} &
                                           0.7432 & \tiny{$\pm$0.0278} &
                                           0.7136 & \tiny{$\pm$0.0131} &
                                            0.7237         & \tiny{$\pm$0.0075} & 
                                            \textit{0.7618}         & \tiny{$\pm$0.0093} \\
                                & Naive WL & \textbf{0.7688} & \tiny{$\pm$0.0060} &
                                             0.6933           & \tiny{$\pm$0.0114} &
                                             \textbf{0.7278}  & \tiny{$\pm$0.0104} &
                                            0.6986         & \tiny{$\pm$0.0115} & 
                                            0.7193        & \tiny{$\pm$0.0110} \\
                                & C-WL     & \textbf{0.7499}  & \tiny{$\pm$0.0056} & 
                                             \textbf{\textit{0.7609}} & \tiny{$\pm$0.0093} & 
                                             \textbf{\textit{0.7684}} & \tiny{$\pm$0.0125} &
                                            \textbf{0.7397} & \tiny{$\pm$0.0124} & 
                                            0.7425        & \tiny{$\pm$0.0076} \\
                                & G-WL     & \textbf{\textit{0.7707}} & \tiny{$\pm$0.0050} & 
                                                      0.7419  & \tiny{$\pm$0.0084} & 
                                             \textbf{0.7578}  & \tiny{$\pm$0.0067} &
                                            \textbf{\textit{0.7584}} & \tiny{$\pm$0.0097} & 
                                            0.7306         & \tiny{$\pm$0.0105} \\
        \hline
        \multirow{4}{*}{HIV} & Atomic & 0.7507 & \tiny{$\pm$0.0199} &
                                        0.7037 & \tiny{$\pm$0.0151} &
                                        0.7044 & \tiny{$\pm$0.0169} &
                                            0.7108         & \tiny{$\pm$0.0075} & 
                                            0.6533         & \tiny{$\pm$0.0279} \\
                             & Naive WL &          0.7286 & \tiny{$\pm$0.0164} &
                                                   0.6939 & \tiny{$\pm$0.0222} &
                                          \textbf{\textit{0.7186}} & \tiny{$\pm$0.0210} &
                                            0.7044         & \tiny{$\pm$0.0115} & 
                                            \textbf{\textit{0.7521}} & \tiny{$\pm$0.0082} \\
                             & C-WL     & \textbf{\textit{0.7545}} & \tiny{$\pm$0.0073} & 
                                          \textbf{\textit{0.7243}} & \tiny{$\pm$0.0243} & 
                                          \textbf{0.7183} & \tiny{$\pm$0.0196} &
                                            \textbf{\textit{0.7242}}  & \tiny{$\pm$0.0141} & 
                                            \textbf{0.7126} & \tiny{$\pm$0.0076} \\
                             & G-WL     &          0.7380 & \tiny{$\pm$0.0164} & 
                                          \textbf{0.7117}& \tiny{$\pm$0.0185} & 
                                          \textbf{0.7131 }& \tiny{$\pm$0.0132} &
                                            0.7038         & \tiny{$\pm$0.0167} & 
                                            \textbf{0.7517} & \tiny{$\pm$0.0081} \\
        \hline
        \multirow{4}{*}{ToxCast} & Atomic & 0.7745 & \tiny{$\pm$0.0083} &
                                            0.7820 & \tiny{$\pm$0.0054} &
                                            0.7720 & \tiny{$\pm$0.0110} &
                                            0.7691         & \tiny{$\pm$0.0066} & 
                                            0.7585         & \tiny{$\pm$0.0108} \\
                                 & Naive WL & \textbf{\textit{0.7927}}& \tiny{$\pm$0.0062} &
                                              0.7462          & \tiny{$\pm$0.0113} &
                                              0.7498          & \tiny{$\pm$0.0093} &
                                            \textbf{0.7734} & \tiny{$\pm$0.0112} & 
                                            0.7580       & \tiny{$\pm$0.0074} \\
                                 & C-WL     & \textbf{0.7831} & \tiny{$\pm$0.0041} & 
                                              \textbf{\textit{0.8013}}& \tiny{$\pm$0.0148} & 
                                              \textbf{0.7754} & \tiny{$\pm$0.0007} &
                                            \textbf{\textit{0.7794}} & \tiny{$\pm$0.0026} & 
                                            \textbf{\textit{0.7708}} & \tiny{$\pm$0.0071} \\
                                 & G-WL     & \textbf{0.7788} & \tiny{$\pm$0.0106} & 
                                                       0.7803 & \tiny{$\pm$0.0002} & 
                                             \textbf{\textit{0.7875}} & \tiny{$\pm$0.0111} &
                                            \textbf{0.7744}  & \tiny{$\pm$0.0085} & 
                                            0.7301        & \tiny{$\pm$0.0083} \\
     \hline
        \multirow{4}{*}{clintox} & Atomic & 0.9387 & \tiny{$\pm$0.0071} &
                                            0.9287 & \tiny{$\pm$0.0153} &
                                            0.9338 & \tiny{$\pm$0.0115} &
                                            0.9305 & \tiny{$\pm$0.0092} &
                                            0.9214 & \tiny{$\pm$0.0153} \\
                                 & Naive WL & 0.9370 & \tiny{$\pm$0.0044} &
                                              \textbf{0.9358} & \tiny{$\pm$0.0001} &
                                              \textbf{0.9429} & \tiny{$\pm$0.0119} &
                                            0.9255 & \tiny{$\pm$0.0060} & 
                                            \textbf{0.9336} & \tiny{$\pm$0.0074} \\
                                 & C-WL & \textbf{\textit{0.9545}} & \tiny{$\pm$0.0054} & 
                                          \textbf{\textit{0.9359}} & \tiny{$\pm$0.0003} & 
                                          \textbf{0.9438} & \tiny{$\pm$0.0078} &
                                          \textbf{\textit{0.9331}} & \tiny{$\pm$0.0142} & 
                                          \textbf{\textit{0.9409}} & \tiny{$\pm$0.0072} \\
                                 & G-WL & 0.9353 & \tiny{$\pm$0.0015} & 
                                        0.9205 & \tiny{$\pm$0.0299} & 
                                        \textbf{\textit{0.9511}} & \tiny{$\pm$0.0062} &
                                        0.9291 & \tiny{$\pm$0.0131} & 
                                        \textbf{0.9358} & \tiny{$\pm$0.0094} \\
        \hline
        \multirow{4}{*}{muv} & Atomic & 0.6350 & \tiny{$\pm$0.0159} &
                                        0.6164 & \tiny{$\pm$0.0907} &
                                        0.6586 & \tiny{$\pm$0.0295} &
                                        0.5716 & \tiny{$\pm$0.0231} & 
                                        0.6839 & \tiny{$\pm$0.0405} \\
                                 & Naive WL & \textbf{\textit{0.7128}}& \tiny{$\pm$0.0214} &
                                        \textbf{\textit{0.6724}} & \tiny{$\pm$0.02265} &
                                        \textbf{0.7324} & \tiny{$\pm$0.0178} &
                                        \textbf{\textit{0.6502}} & \tiny{$\pm$0.01792} & 
                                        \textbf{\textit{0.7362}} & \tiny{$\pm$0.0195} \\
                                 & C-WL & \textbf{0.6853} & \tiny{$\pm$0.0192} & 
                                    \textbf{0.6677} & \tiny{$\pm$0.0281} & 
                                    \textbf{\textit{0.7517}} & \tiny{$\pm$0.0007} &
                                    \textbf{0.5771} & \tiny{$\pm$0.0217} & 
                                    \textbf{0.7127} & \tiny{$\pm$0.0192} \\
                                 & G-WL & \textbf{0.6806} & \tiny{$\pm$0.0172} & 
                                    \textbf{0.6387} & \tiny{$\pm$0.0326} & 
                                    \textbf{0.6912} & \tiny{$\pm$0.0406} &
                                    \textbf{0.6003} & \tiny{$\pm$0.0392} & 
                                    \textbf{0.7117} & \tiny{$\pm$0.0192} 
          \end{tabular}   
    \caption{Ten-run averages of ROC-AUCs for the classification benchmarks. Larger values are better. Bold indicates that the WL embedding improves the generalization performance from atomic embedding. Italics indicate the best embedding within each cell.  }
    \label{tab:optimized_result_classification_full}
\end{table*}

\section{Additional Experiment Result}

\subsection{Empirical assessment of difficulty of recoverng WLE with atomic embedding + GNN layer}

In order to support our theoretical analysis in Sec.4, we empirically study whether one non-linear GNN layer (with a large unit size) plus the atomic embedding can extract the information similar to the WLE. 

In this experiment, the atomic embedding is followed by \textbf{a single layer of nonlinear GNN (more specifically, GCN~\citep{Kipf_Welling17ICLR})} and the readout layer. 
In contrast, we apply \textbf{no GNN layers after the three WLEs} (naive, C-WL, and G-WL): we directly apply the readout to the WL embedding vectors. 

\begin{figure}
    \centering
    \includegraphics[width=80mm]{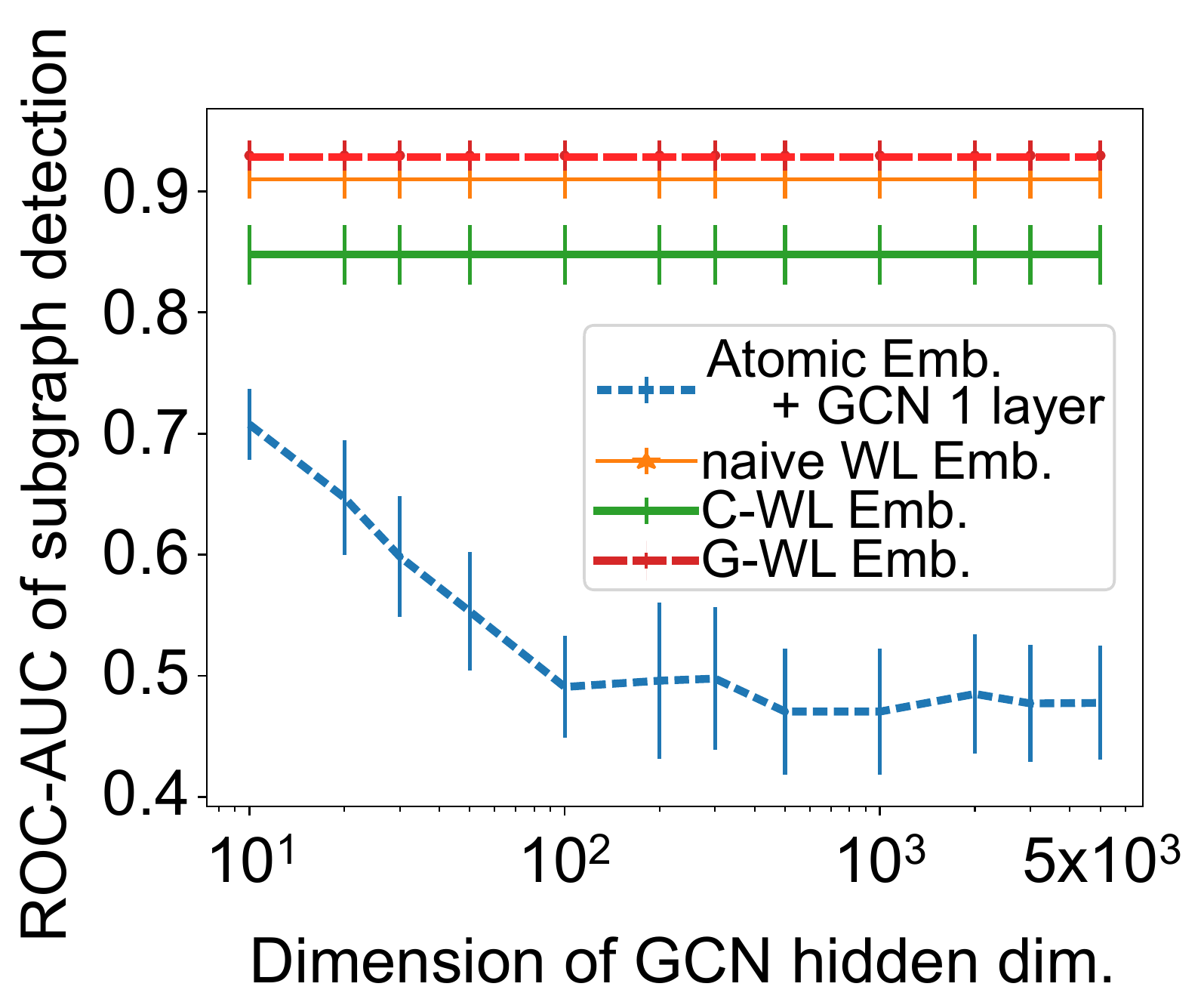}
    \caption{Subgraph Detection experiments, Atomic embedding + single GCN VS. WL Embedding. AUC evolution over hidden dim. of a single GNN layer. }
    \label{fig:subgraph_detection_vs_dims}
\end{figure}

The ROC-AUC evolution over the dimension of the GCN hidden vectors (for atomic embedding) is presented in Fig.~\ref{fig:subgraph_detection_vs_dims}. 
As the figure shows, the GCN with atomic embedding never achieves the AUCs comparable to the WLEs, regardless of the hidden vectors' dimension. 
The evolution of the training losses are presented in Fig.~\ref{fig:subgraph_detection_vs_dims_loss} of the main manuscript. 
It is notable that the atomic embedding network cannot decrease the training loss as small as those of WLEs, regardless of the model complexity (hidden dimensions). 
These results indicate that it is very difficult to emulate WLE-like features by typical MPNNs plus atomic embeddings. 

\subsection{Analysis of Learned WL Embedding}

In this section, we investigate how WL embedding affects to the representation learning in GNNs. As a particular example, we pick up the combination of C-WL embedding and GCN. 

\subsubsection{Visualization of Learned Weight}
We examine the contributions of the atomic embedding vectors $z_{i,\ell}$ and the neighboring embedding vectors $z_{i,\mathcal{M}}$ in Eq.~\ref{eq:CWL_embed} based on the weight matrix $W$, which we expect to indicate the importance of the embedding vectors by its scale. The left half of $W$ corresponds to $z_{i,\ell}$ and the right half of $W$ corresponds to $z_{i,\mathcal{M}}$. To eliminate the scale invariance between the embedding vectors and the weight, we normalize $z_{i,\ell}$ and $z_{i,\mathcal{M}}$ and rescale $W$.

\begin{figure}[t]
    \centering
    \includegraphics[width=80mm]{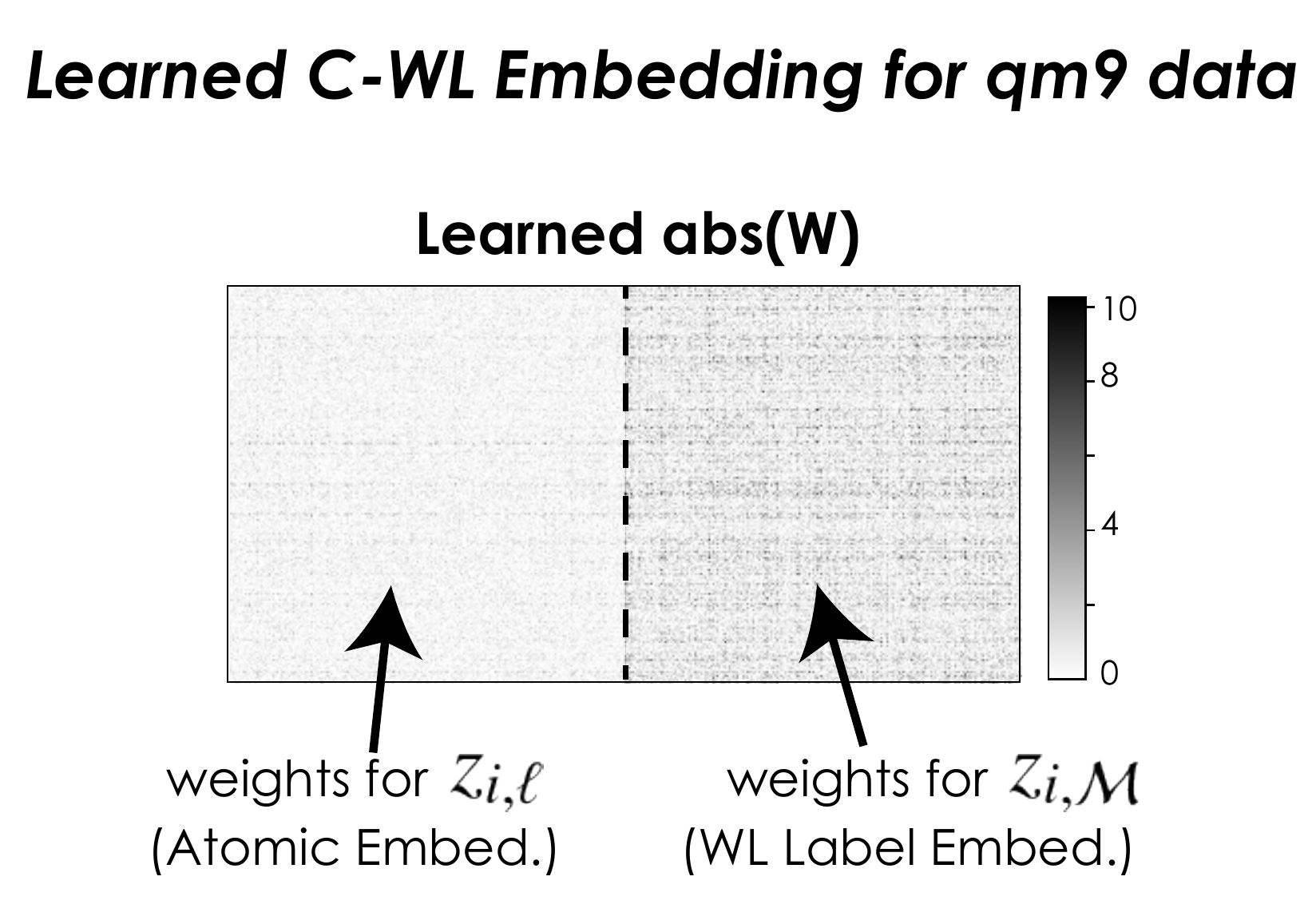}
    \includegraphics[width=80mm]{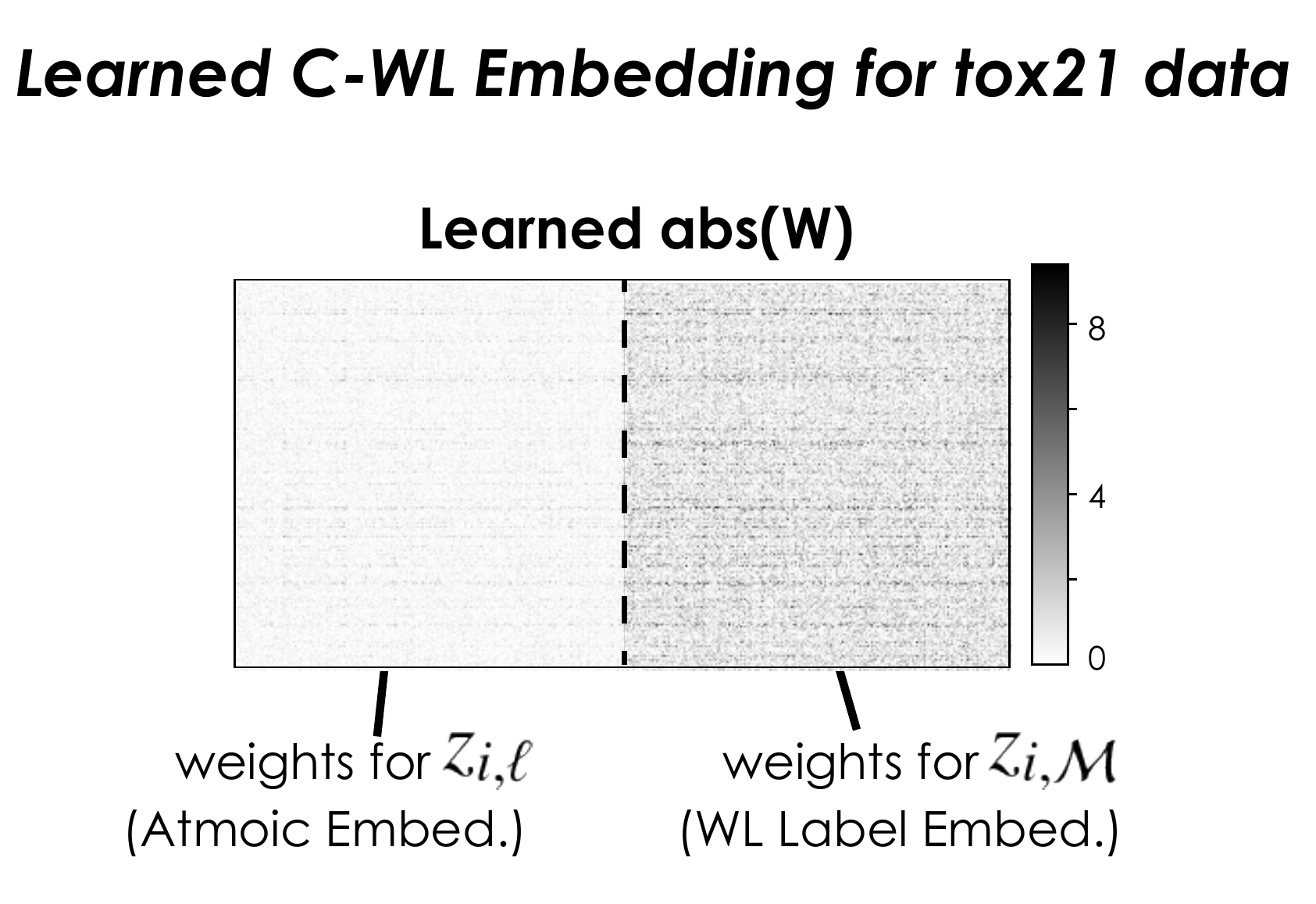}
    \caption{Visualization of the learned weight matrix $W$ of the C-WL Embedding for the QM9 dataset (upper panel) and the Tox21 dataset (lower panel). The absolute values of matrix entries are presented. }
    \label{fig:CWL_qm9_tox21_W}
\end{figure}

Fig.~\ref{fig:CWL_qm9_tox21_W} visualizes the absolute values of $W$ trained with the QM9 and Tox21 datasets, with which WL embedding made large improvements. The visualization clearly depicts that the weight element for $z_{i,\mathcal{M}}$ is more active than that for $z_{i,\ell}$.

\begin{figure}[t]
    \centering
    \includegraphics[width=80mm]{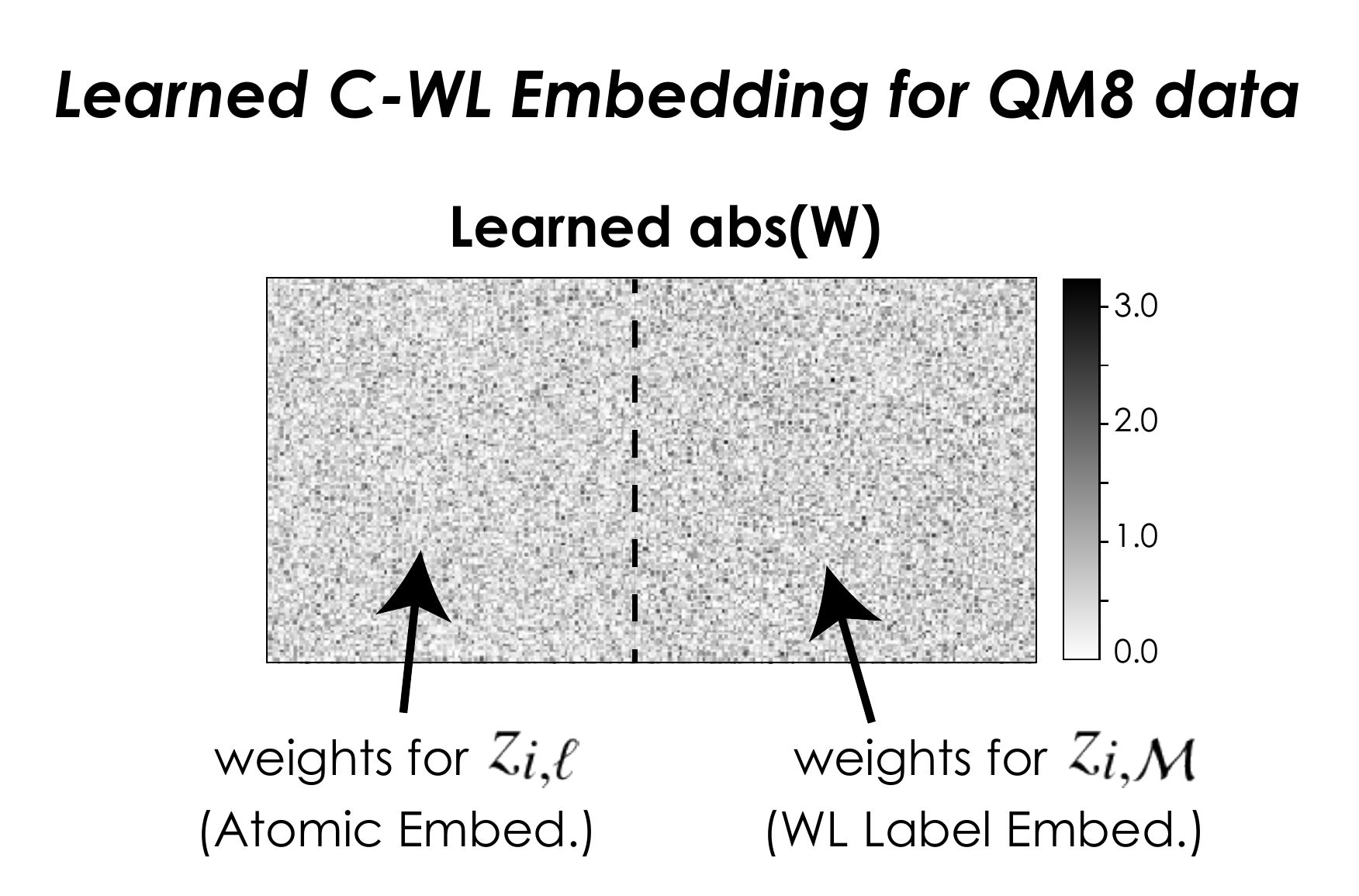}
    \includegraphics[width=80mm]{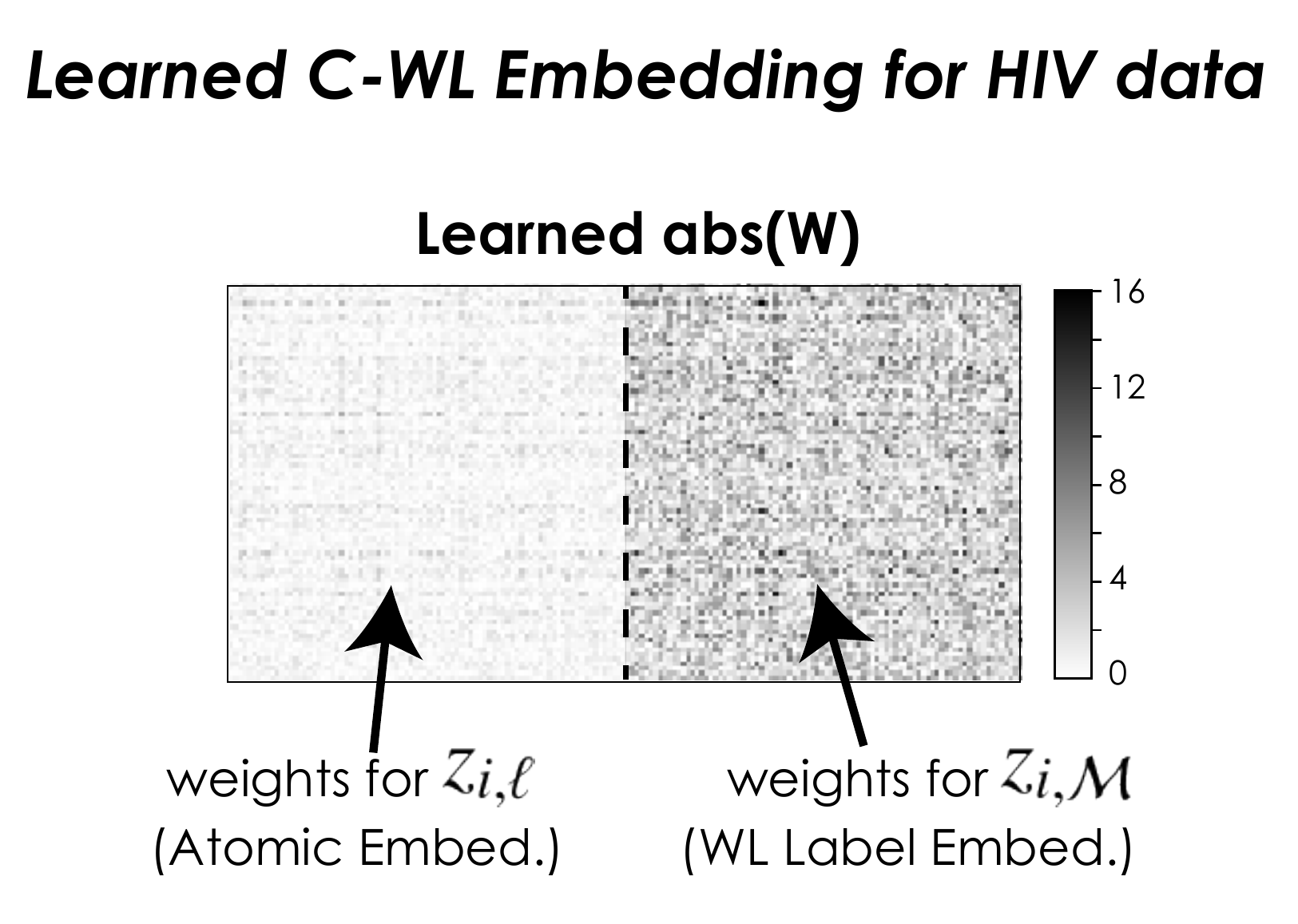}
    \caption{Visualization of the learned weight matrix $W$ of the C-WL Embedding for the QM8 dataset (upper panel) and the HIV dataset (lower panel). The absolute values of matrix entries are presented.  The C-WL embedding records small improvements in these datasets. }
    \label{fig:CWL_qm8_HIV_W}
\end{figure}

Fig.~\ref{fig:CWL_qm8_HIV_W} shows the weight matrix of the QM8 dataset and the HIV dataset, where we observe smaller performance improvements by C-WL embedding. 
We note that the C-WL embedding never zero-suppress weights for $\mathcal{M}$, even if we do not achieve large improvements in downstream task performance. The embedding tries to find a good balance between original atom labels and neighboring labels, and this contributes slightly to the downstream tasks. 

\subsubsection{Importance of Neighboring Labels}

Next, we quantitatively analyze the importance of neighboring labels. We employ node label shuffling inspired by the permutation (feature) importance~\citep{Breiman01}. First, we train GNNs with the C-WL embedding in a usual manner.
After training, we randomly choose a node $i$ for each sample in a test dataset and change its node label $\ell_i$ or neighboring label $\mathcal{M}_i$ to a different label. Therefore, if some specific labels are critically important, the shuffling of them will seriously degrade the prediction performance.

We manipulate the label in two ways: \textit{Atom shuffle} and \textit{Neighboring Label (NL) shuffle}.
The Atom shuffle replaces the node label $\ell_{i}$ by the pseudo one $\widehat{\ell}_{i}$, while the NL shuffle replaces the neighbor label $\mathcal{M}_i$ with $\widehat{\mathcal{M}}_i$.
Formally, we overwrite embeddings as follows:
\begin{align}
    \text{Atom Shuffle} \quad & z_{i, \ell}\leftarrow \text{WLEmbed}_{1} (\widehat{\ell}_{i}, \emptyset)\\
    \text{NL Shuffle} \quad & z_{i, \mathcal{M}}\leftarrow \text{WLEmbed}_{2} (\emptyset, \widehat{\mathcal{M}}_{i}).
\end{align}
We compute pseudo labels $\widehat{\ell}_{i}$ and $\widehat{\mathcal{M}}_{i}$ 
so that they are different from $\ell_{i}$ and $\mathcal{M}_{i}$, respectively.

From the results (Table~\ref{tab:PI}), we observe that the NL Shuffle has a larger impact on performances than the Atom shuffle in all datasets.
It implies that the neighboring labels $\mathcal{M}$ are considered as informative cues for molecule property predictions in WL embedding-based GNNs.

\begin{table}[t]
    \centering
    \begin{tabular}{c||ccc}
        Dataset & No Shuffle & Atom Shuffle & NL Shuffle \\ \hline
        LIPO & 0.6524& 0.6718 & \textbf{0.6952}\\
        QM9& 10.4956& 17.2870 & \textbf{17.7292}\\
        QM8 & 0.0306& 0.0329  & \textbf{0.0335} \\\hline
        Tox21 & 0.7499 & 0.7413& \textbf{0.6637} \\
        HIV & 0.7545 & 0.7457 & \textbf{0.7059} \\
        Toxcast & 0.7831& 0.7819& \textbf{0.7798} \\
    \end{tabular}
    \caption{10-run averages MAE (upper) and ROC-AUC (lower) on shuffled test data. ``No shuffle'' results are borrowed Tables \ref{tab:optimized_result_regression} and \ref{tab:optimized_result_classification}.
    Bold numbers indicate the \textit{worst} performances.}
    \label{tab:PI}
\end{table}

\subsection{Iterative Label Expansion}

We empirically evaluate how the iterative label expansion discussed in Section \ref{sec:iterative_label_expansion} affects the model performance.
Table \ref{tab:multi_label_expansion_qm9_full} shows the performance of models with the number of label expansion iterations $T$ varying from $1$ to $3$.
We use a GCN as a base GNN.
All experiment settings are the same as described in Section~\ref{sec:experiment_setting}, except that we conduct a hyperparameter optimization for every run. 
We present the result for the QM9, QM8, and the Tox21 dataset. 

The results for the QM9 dataset (Table~\ref{tab:multi_label_expansion_qm9_full}) indicate a trade-off of the representation power and model complexity. For the naive WL embedding, we observe that a performance steadily degrades while increasing the number of iterations $T$.
Notably, the $T \! = \! 3$ case is significantly worse than that of $T \! = \! 2$. The increased number of expanded labels $J$ makes the input feature vectors significantly sparse when $T \! = \! 3$ (for the QM9 dataset, $J \!= \!4958$ for $T \!= \! 2$ while $J \! = \! 168225$ for $T \! = \! 3$). 

For C-WL and G-WL embedding, $T \! = \! 2$ works better than $T \! = \! 1$.  
As described in Section \ref{sec:c_wl_g_wl}, naive WL embedding maps different expanded labels at completely different points even if their original labels are the same. This does not hold for the C-WL and the G-WL embedding thanks to the separated $z_{i,\ell}$ and $z_{i, \mathcal{M}}$. 
Therefore, we hypothesize that C-WL and G-WL are more robust to the explosion of extended label patterns than naive WL.
In addition, we can infer that the large sample size and the complexity of the QM9 dataset may prevent C-WL and G-WL embedding with $T \! = \! 2$ from overfitting.
In fact, we observe that $T \! = \! 1$ achieves better results for C-WL and G-WL for the QM8 dataset (Table~\ref{tab:multi_label_expansion_qm8_full}), which has a smaller sample size.

\begin{table}[t]
    \centering
    \begin{tabular}{cc||rr@{\hskip 0.5\tabcolsep}c@{\hskip 0.5\tabcolsep}r}
        Embedding & $T$ & $J$ & \multicolumn{3}{c}{MAE}\\ \hline
        Atomic      & --- &      10 & 17.0581&$\pm$&1.3320 \\
        Naive WL  & $1$ &      49 &  \textbf{8.6073} &$\pm$& \textbf{0.3279} \\
                  & $2$ &   4,958 &  8.8921 &$\pm$& 0.1944 \\
                  & $3$ & 168,225 & 14.3417 &$\pm$& 0.3956 \\
        C-WL      & $1$ &      31 &  8.5097 &$\pm$& 0.2675 \\
                  & $2$ &   4,826 &  \textbf{8.2116} &$\pm$& \textbf{0.2752} \\
                  & $3$ & 160,758 &  \multicolumn{3}{c}{N.A.}\\ 
        G-WL      & $1$ &      31 &  8.2913 &$\pm$& 0.2411 \\
                  & $2$ &   4,826 &  \textbf{*7.9231} &$\pm$& \textbf{0.1276} \\
                  & $3$ & 160,758 &  \multicolumn{3}{c}{N.A.}
    \end{tabular}
    \caption{
    The number of expanded labels and MAE of the embeddings with iterative label expansions for the QM9 dataset. $T$ and $J$ denote the number of iterations and the number of (expanded) labels, respectively. We cannot obtain results due to GPU errors for C-WL and G-WL with $T=3$ (``N.A.") . Bold numbers indicate the best models within each model and the asterisk the best model among all configurations.
    }
    \label{tab:multi_label_expansion_qm9_full}
\end{table}

\begin{table}[t]
    \centering
    \begin{tabular}{cc||rr@{\hskip 0.5\tabcolsep}c@{\hskip 0.5\tabcolsep}r}
        Model       & $T$ & $J$ & \multicolumn{3}{c}{MAE}  \\ \hline
        Atomic & --- &           10 & $0.02987$ &$\pm$& $0.0001$ \\
        Naive WL   & $1$ &     49 & \textbf{*0.02866} &$\pm$& \textbf{0.0004} \\
                       & $2$ &  2,968 & $0.02973$ &$\pm$& $0.0002$ \\
                       & $3$ & 39,002 & $0.03363$ &$\pm$& $0.0005$ \\
        C-WL     & $1$ &     31 & \textbf{0.02927} &$\pm$& \textbf{0.0002} \\
                       & $2$ &  2,865 & $0.02959$& $\pm$& $0.0003$ \\
                       & $3$ & 38,570 & \multicolumn{3}{c}{N.A.}\\
        G-WL     & $1$ &     31 & \textbf{0.02874} &$\pm$& \textbf{0.0002} \\
                       & $2$ &  2,865 & $0.02940$& $\pm$& $0.0002$ \\
                       & $3$ & 38,570 & \multicolumn{3}{c}{N.A.}
    \end{tabular}
    \caption{The number of expanded labels and MAE of the embeddings with iterative label expansions for the QM8 dataset. Symbols are same as those of Table \ref{tab:multi_label_expansion_qm9_full}.}    
    \label{tab:multi_label_expansion_qm8_full}
\end{table}

Table~\ref{tab:multi_label_expansion_tox21_full} shows the model performances, including correct G-WL, using the Tox21 dataset for various of number of iterations.

\begin{table}[t]
    \centering
    \begin{tabular}{cc||rr@{\hskip 0.5\tabcolsep}c@{\hskip 0.5\tabcolsep}r}
        Model                   & $T$ & $J$ & \multicolumn{3}{c}{ROC-AUC}  \\ \hline
        Atomic & --- &           84 & $0.7429$& $\pm$& $0.0047$ \\
        Naive  WL   & $1$ &    457 & \textbf{0.7724} &$\pm$& \textbf{0.0047} \\
                      & $2$ &  4,695 & $0.7473$& $\pm$& $0.0031$ \\
                      & $3$ & 25,207 & $0.7211$& $\pm$& $0.0032$ \\
        C-WL     & $1$ &    237 & \textbf{0.7714} &$\pm$& \textbf{0.0025} \\
                      & $2$ &  4,500 & \multicolumn{3}{c}{N.A.} \\
                      & $3$ & 24,916 & \multicolumn{3}{c}{N.A.}\\
        G-WL     & $1$ &    237 & \textbf{*0.7744} &$\pm$& \textbf{0.0026} \\
                      & $2$ &  4,500 & \multicolumn{3}{c}{N.A.} \\
                      & $3$ & 24,916 & \multicolumn{3}{c}{N.A.}
    \end{tabular}
    \caption{The number of expanded labels and ROC-AUC of the embeddings with iterative label expansions for the Tox21 dataset. Symbols are same as those of Table \ref{tab:multi_label_expansion_qm9_full}.}
    \label{tab:multi_label_expansion_tox21_full}
\end{table}

\end{document}